\documentclass[10pt]{report}
\usepackage[utf8]{inputenc}
\usepackage[T1]{fontenc}

\usepackage{amsmath, amsthm, amsfonts, amssymb}
\usepackage{mathtools}

\usepackage{graphicx}
\usepackage{graphics}
\usepackage{epstopdf}
\usepackage{psfrag}
\usepackage{rotating}
\usepackage{xcolor}

\usepackage[left=2cm, right=2cm, top=1.5cm, bottom=2cm]{geometry}

\usepackage{multirow, caption, subcaption, tabularx, booktabs}
\usepackage{array}
\usepackage{longtable}
\usepackage{threeparttable}
\usepackage{colortbl}

\let\algorithm\relax
\let\algorithm*\relax

\usepackage{algorithm}
\usepackage{algpseudocode}

\usepackage{moreverb}
\usepackage{listings}

\definecolor{jsonkey}{rgb}{0.5,0,0}
\definecolor{jsonstring}{rgb}{0,0,0.6}

\lstdefinelanguage{json}{
    basicstyle=\ttfamily\small,
    numbers=left,
    numberstyle=\tiny\color{gray},
    stepnumber=1,
    numbersep=5pt,
    showstringspaces=false,
    breaklines=true,
    frame=single,
    backgroundcolor=\color{white},
    literate=
     *{0}{{{\color{jsonstring}0}}}{1}
      {1}{{{\color{jsonstring}1}}}{1}
      {2}{{{\color{jsonstring}2}}}{1}
      {3}{{{\color{jsonstring}3}}}{1}
      {4}{{{\color{jsonstring}4}}}{1}
      {5}{{{\color{jsonstring}5}}}{1}
      {6}{{{\color{jsonstring}6}}}{1}
      {7}{{{\color{jsonstring}7}}}{1}
      {8}{{{\color{jsonstring}8}}}{1}
      {9}{{{\color{jsonstring}9}}}{1}
      {:}{{{\color{black}:}}}{1}
      {,}{{{\color{black},}}}{1}
      {"}{{{\color{black}"}}}{1},
    morestring=[b]",
    stringstyle=\color{jsonstring},
    keywordstyle=\color{jsonkey},
}

\usepackage{tikz}
\usetikzlibrary{positioning, arrows.meta, shadows, fit}
\usepackage[all]{xy}

\usepackage[below]{placeins}
\usepackage{float}

\usepackage{lmodern}
\usepackage{url}
\usepackage{hyperref}
\usepackage{cleveref}

\usepackage{authblk}
\usepackage{titlesec}

\usepackage{framed}
\usepackage{enumitem}
\usepackage{lineno}
\usepackage[round]{natbib}

\titleformat{\section}{\bfseries\large}{\thesection.}{1em}{}
\titleformat{\subsection}{\bfseries}{\thesubsection.}{1em}{}

\swapnumbers  
\theoremstyle{plain}

\theoremstyle{remark}

\theoremstyle{definition}

\numberwithin{equation}{section}

\title{Exploring Large Language Models for Financial Applications: Techniques, Performance, and Challenges with FinMA}

\author[1,2]{P. Djagba\thanks{djagbapr@msu.edu}}
\author[4]{A. Younoussi Saley\thanks{saley.younoussi@aims.ac.rw}}

\affil[1]{Lyman Briggs College, Michigan State University}
\affil[2]{Department of Finance, Michigan State University}
\affil[4]{African Institute for Mathematical Sciences, Rwanda}

\date{}

\begin{document}

\maketitle

\begin{abstract}
This research explores the strengths and weaknesses of domain-adapted Large Language Models (LLMs) in the context of financial natural language processing (NLP). The analysis centers on FinMA, a model created within the PIXIU framework, which is evaluated for its performance in specialized financial tasks. Recognizing the critical demands of accuracy, reliability, and domain adaptation in financial applications, this study examines FinMA's model architecture, its instruction tuning process utilizing the Financial Instruction Tuning (FIT) dataset, and its evaluation under the FLARE benchmark. Findings indicate that FinMA performs well in sentiment analysis and classification, but faces notable challenges in tasks involving numerical reasoning, entity recognition, and summarization. This work aims to advance the understanding of how financial LLMs can be effectively designed and evaluated to assist in finance-related decision-making processes.\\

\end{abstract}

\tableofcontents
\newpage

\section{Introduction}

\section{Context}

Large Language Models (LLMs) have significantly influenced the advancement of Natural Language Processing (NLP), exhibiting strong performance across diverse linguistic tasks \cite{Lee2024survey, Touvron2023llama}. The evolution of LLMs was notably accelerated since the introduction of the Transformer architecture by \cite{Vaswani2017transformer}, which replaced recurrent structures with self-attention mechanisms. This innovation enhanced model scalability, parallel processing capabilities, and training efficiency, enabling the development of powerful models such as ChatGPT (Openai2022) until GPT-4 \cite{openai2024gpt4technicalreport} now.

These breakthroughs have opened new research frontiers across various disciplines, including mathematics, science, healthcare, and finance \cite{Chen2024}. Within the financial sector, the inherent complexity, rapid fluctuations, and information asymmetries present substantial analytical challenges. LLMs have emerged as valuable tools due to their capacity to comprehend contextual nuances, process extensive unstructured data, and produce coherent, human like outputs \cite{Nie2024}.

Given the domain-specific nature of financial language, which involves highly technical and context-dependent terminology, several LLMs have been developed to address these unique characteristics. Early efforts include FinBERT \cite{Araci2019finbert}, a model adapted from BERT\cite{devlin2019bertpretrainingdeepbidirectional} for sentiment analysis in financial texts. Subsequent developments such as BloombergGPT \cite{wu2023bloomberggpt}, FinGPT \cite{Lee2024fingpt}, and FinMA \cite{Xie2023pixiu} reflect a progression toward more sophisticated architectures tailored for financial applications, incorporating structured knowledge and large-scale web data.

These specialized models support a variety of financial NLP tasks, including sentiment analysis, named entity recognition, text classification, question answering, or stock trend forecasting. Their applications are instrumental in automating information extraction and enabling advanced analytics such as trading strategy formulation and risk evaluation \cite{Lee2024survey, Nie2024}.

The progress in this area underscores the necessity of designing LLMs that are finely tuned to financial domains and tasks, particularly where timing, accuracy, and interpretability are paramount.

\section{Problem Statement}
\label{sec:problem}
While general-domain LLMs have been extensively studied, Financial LLMs (FinLLMs) remain an emerging area with limited comprehensive surveys \cite{Lee2024survey}, and several challenges persist. The domain faces significant data accessibility constraints, where high-quality, specialized financial datasets remain largely proprietary or possess insufficient coverage for comprehensive model training \cite{Nie2024}. Financial natural language processing introduces heightened complexity through tasks requiring advanced numerical reasoning and causal inference capabilities that extend beyond conventional text generation paradigms \cite{Lee2024survey}. Current evaluation frameworks present limitations in adequately assessing FinLLM performance, lacking specialized benchmarks that accurately reflect practical financial application requirements rather than general linguistic competency \cite{Lee2024survey, Chen2024}. Additionally, the financial sector imposes stringent reliability standards, necessitating exceptional accuracy, interpretability, and dependability given the substantial economic consequences of erroneous predictions or misinterpretations \cite{Lee2024survey}. These multifaceted challenges underscore the critical need for comprehensive analysis of existing FinLLM architectures, training approaches, evaluation methodologies, and application domains to inform future research directions in this rapidly advancing field.

\section{Objectives}
\label{sec:objectives}

The primary aim of this study is to examine the strengths and limitations of domain-specific Large Language Models (FinLLMs) in executing financial natural language processing (NLP) tasks, with particular emphasis on the FinMA model \cite{Xie2023pixiu}.

This research is guided by the following specific objectives:

\begin{itemize}
    \item To investigate the architectural and training distinctions between general-purpose LLMs and those designed for financial applications.
    \item To assess the performance of FinMA-7B-full on benchmark financial tasks and datasets, using established models such as BloombergGPT and GPT-4 as comparative baselines.
    \item To identify the key challenges and opportunities associated with the development and deployment of FinLLMs in real-world financial settings.
\end{itemize}

To support these objectives, this research addresses the following questions:
\begin{itemize}
    \item What architectural adaptations and training methodologies enable LLMs to effectively process and interpret financial information?
    \item How does FinMA compare with general-purpose models in financial reasoning, prediction tasks, and specialized applications?
    \item What evaluation frameworks most effectively measure LLM performance in financial contexts?
\end{itemize}

\section{Methodology}
\label{sec:methodology}

This thesis addresses the identified gaps by conducting an analysis of FinMA, a prominent financial large language model from the PIXIU framework \cite{Xie2023pixiu}. The model is selected due to its open-source availability and relevance to financial NLP tasks. Datasets and evaluation tools are sourced from the Hugging Face platform, ensuring accessibility and reproducibility.

The research employs a mixed-methods approach that integrates both quantitative and qualitative techniques to address the research questions. It begins with a systematic literature review that traces the development of financial large language models (FinLLMs), examining their architectural evolution, training methodologies, and evaluation benchmarks as documented in existing studies. This is followed by an empirical analysis that assesses the performance of FinMA in comparison to general-purpose models such as GPT-4 across a range of financial natural language processing (NLP) tasks, using established benchmark datasets. Finally, a qualitative discussion is conducted to interpret the evaluation outcomes, highlighting key challenges such as hallucination and data privacy, as well as emerging opportunities, including the potential for multimodal integration. A detailed account of the methodology, encompassing the FinMA architecture, the datasets utilized, and the evaluation framework, is presented in Chapter~\ref{chap:methodology}.

\section{Thesis Structure}

The rest of this thesis is organized as follows:
\begin{itemize}
    \item \textbf{Chapter 2: Literature Review} — This chapter reviews past research on LLMs, financial NLP applications, and specialized language models. It builds the literature background needed for this study.
    \item \textbf{Chapter 3: Methodology} — This chapter explains the research design, evaluation methods, and analysis techniques used. It describes FinMA’s architecture, training methods, and evaluation framework.
    \item \textbf{Chapter 4: Results and Analysis} — This chapter presents the experimental setup, compares FinMA’s performance with baselines like BloombergGPT and GPT-4, and discusses challenges and solutions.
    \item \textbf{Chapter 5: Conclusion and Future Directions} — This chapter summarizes the main findings and suggests ideas for future research.
\end{itemize}

\chapter{Literature Review}
\label{chap:literature_review}

The emergence of Financial Large Language Models (FinLLMs) corresponds with the broader evolution of transformer-based language models, which began with the introduction of the Transformer architecture by \cite{Vaswani2017transformer}. While general-purpose models like BERT \cite{devlin2019bertpretrainingdeepbidirectional} and GPT \cite{Radford2018ImprovingLU} laid the foundation for language understanding, their application in financial contexts revealed performance limitations due to a lack of domain-specific adaptation. Financial texts often contain terminology and syntactic structures that differ significantly from general-language corpora, necessitating the development of specialized FinLLMs \cite{Lee2024survey}.

Early contributions to the field included FinBERT \cite{Araci2019finbert}, which fine-tuned BERT on financial texts to improve sentiment analysis. Subsequent models, such as FLANG \cite{Shah2022flang} and FinGPT \cite{Lee2024fingpt}, expanded these ideas by incorporating larger datasets and instruction tuning. The release of BloombergGPT \cite{wu2023bloomberggpt} demonstrated the potential of large-scale financial pretraining, although its proprietary nature limits reproducibility. FinMA \cite{Xie2023pixiu}, on the other hand, provides an open-source alternative designed to perform across a wide array of financial NLP tasks.

This open-access nature of some FinLLMs contrasts with closed-source models such as BloombergGPT, which restrict access to weights, training data, or model architecture. Open-source models offer transparency, reproducibility, and community collaboration, while closed-source models are typically proprietary and optimized for commercial use.

Recent advancements also emphasize multimodal learning and instruction tuning. Open-FinLLMs \cite{huang2025openfinllms} introduced multimodal variants capable of handling textual, numerical, and tabular data simultaneously. These models signify a shift toward generalist financial agents that can operate across varied data modalities.

This literature review examines the evolution of FinLLMs, their limitations compared to general LLMs, and the current state of the field. It explores the architectures, training methods, and performance benchmarks of notable FinLLMs.

\section{Evolution of Financial Language Models}

The development of financial language models has progressed through several distinct phases since the late 2010s. Initially, researchers focused on adapting existing transformer architectures for financial text processing, leading to the creation of specialized models such as FinBERT \cite{Araci2019finbert}. This foundational work was subsequently extended through iterations including FinBERT-20 \cite{yang2020finbertpretrainedlanguagemodel} and FinBERT-21 (Liu2021finbert21), each building upon previous architectures while incorporating domain-specific improvements.
The field saw methodological innovation with FLANG \cite{Shah2022flang}, which departed from traditional approaches by implementing ELECTRA-based architectures \cite{clark2020electrapretrainingtextencoders} combined with financial-specific token masking strategies. This period marked a shift toward more sophisticated pre-training techniques tailored to financial vocabularies and contexts.
The landscape evolved significantly by 2023, characterized by the introduction of large-scale, comprehensive financial language models. Notable developments included BloombergGPT \cite{wu2023bloomberggpt}, FinGPT \cite{Lee2024fingpt}, and FinMA \cite{Xie2023pixiu}, each representing advances in model scale and capability. More recently, the field has expanded beyond text-only processing, with frameworks like Open-FinLLMs \cite{huang2025openfinllms} incorporating multimodal functionalities to handle diverse financial data types.

\begin{figure}[H]
    \centering
    \includegraphics[width=1\linewidth]{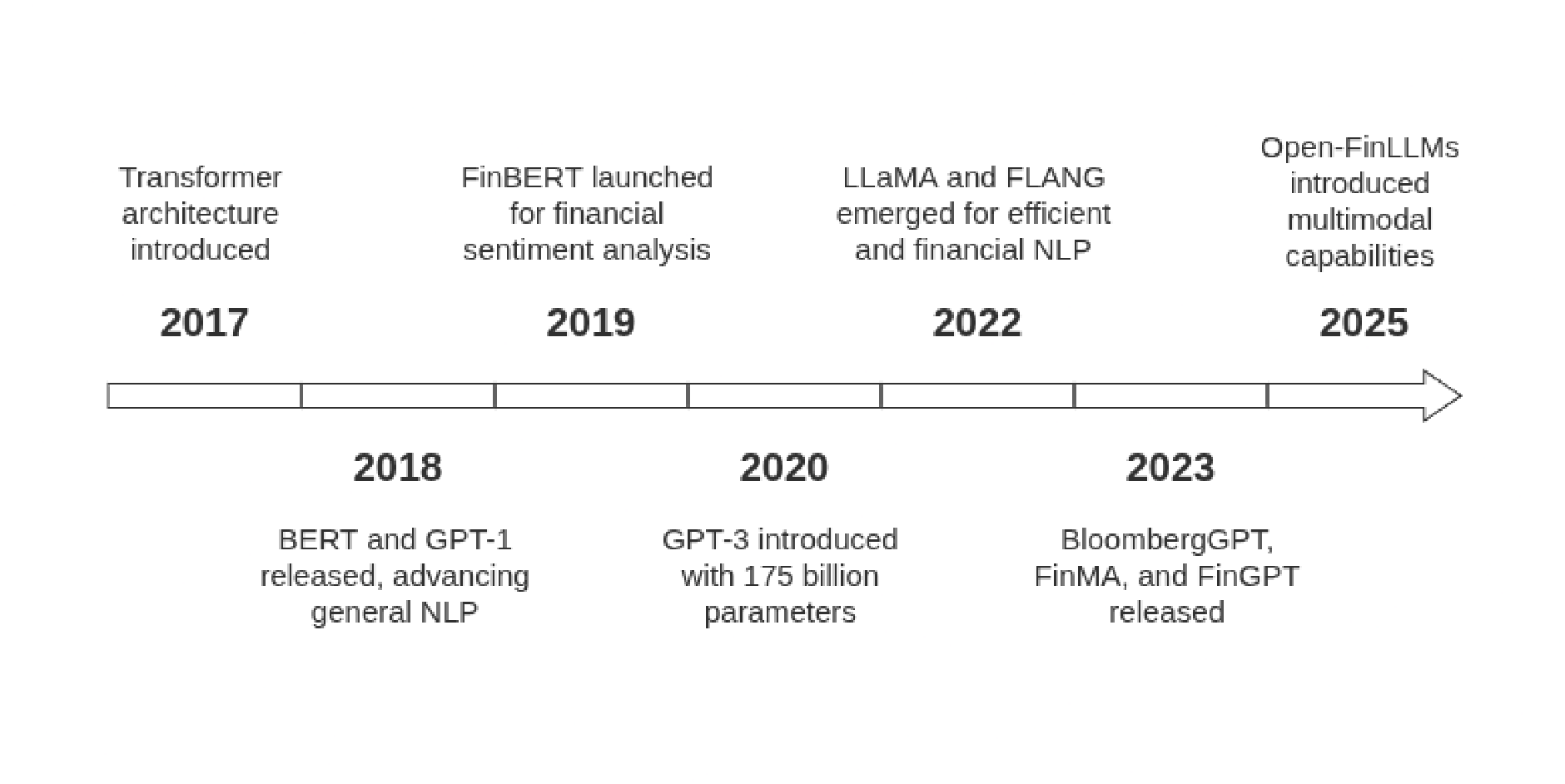}
    \caption{Timeline of key developments in financial natural language processing from 2017 to 2025.}
    \label{fig:keymilstone}
\end{figure}

This evolution is shown in Figure~\ref{fig:timeline}, which illustrates the timeline from general LLMs to specialized FinLLMs. Figure \ref{fig:keymilstone} summurize this key milestones:

\begin{figure}
    \centering
    \includegraphics[width=1\linewidth]{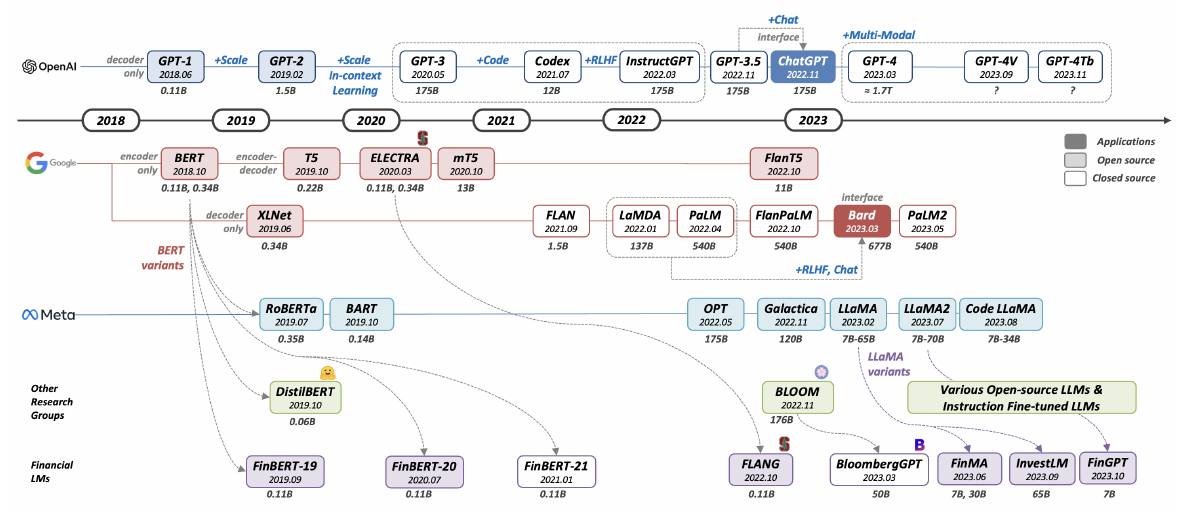}
    \caption{Chronological progression of notable pre-trained language models and large language models from general-purpose applications to finance-specific implementations. [Source: \cite{Lee2024survey}]}
    \label{fig:timeline}
\end{figure}

\section{Limitations of General-Domain LLMs in Finance}
\label{sec:general_limitations}

General-domain LLMs, such as BERT, GPT-3, or GPT-4, are trained on diverse datasets covering a wide range of topics, making them versatile for general language tasks. However, their application in finance is limited by several factors \cite{Nie2024, Chen2024}:
\begin{enumerate}
    \item \textbf{Lack of Domain-Specific Knowledge:} Financial texts contain specialized terminology and concepts that general LLMs may not fully understand. For instance, terms like ``yield'' or ``derivative'' have specific financial meanings that differ from their general usage, leading to potential misinterpretations \cite{Lee2024survey}.
    \item \textbf{Data Efficiency:} General purpose language models require extensive fine-tuning on financial data to achieve competitive performance, which demands significant computational resources. In contrast, finance-specific language models (FinLLMs), which are pretrained or fine-tuned on financial corpora, tend to perform well with less training data, making them more data-efficient for financial tasks \cite{Nie2024}.
    \item \textbf{Cost and Accessibility:} Proprietary general LLMs like BloombergGPT GPT-4 are costly and less accessible, while many FinLLMs, such as FinGPT and FinMA, are open-source, making them more practical for researchers and smaller institutions \cite{Lee2024fingpt}.
\end{enumerate}

While FinLLMs often outperform general purpose models like GPT-4 in specialized financial tasks such as sentiment analysis, numerical reasoning, and regulatory document analysis due to their targeted training on domain specific corpora \cite{Nie2024, Chen2024}, GPT-4 generally achieves superior performance in broader reasoning tasks thanks to its scale, diverse pretraining data, and advanced zero/few-shot capabilities. However, GPT-4 remains a closed source model with limited transparency and deployment flexibility, which poses challenges for financial institutions concerning data privacy, compliance, and customizability. In contrast, several FinLLMs, particularly open source ones such as FinGPT or FinMA, offer greater control, adaptability, and potential for on premise deployment making them better suited for use cases where interpretability and data confidentiality are paramount \cite{Lee2024survey}.

\section{Current Financial Language Models}
\label{sec:current_finllms}

The landscape of FinLLMs includes several models tailored for financial NLP, each with distinct features:
\begin{itemize}
  \item \textbf{FinBERT:} Proposed by Araci, FinBERT is a domain-adapted variant of BERT, pre-trained on financial corpora. It has gained wide usage for sentiment analysis and text classification in finance-related applications~\cite{Araci2019finbert}.

  \item \textbf{FLANG:} Presented by \cite{Shah2022flang}, FLANG is a transformer-based language model optimized for financial text. It achieves strong results in sentiment detection and classification tasks specific to the financial sector~\cite{Shah2022flang}.

  \item \textbf{BloombergGPT:} With 50 billion parameters, BloombergGPT is trained on a hybrid corpus combining general and domain-specific financial texts. It is built to handle a broad spectrum of financial NLP tasks~\cite{wu2023bloomberggpt}.

  \item \textbf{FinGPT:} Introduced by \cite{Lee2024fingpt}, FinGPT is an open-source initiative focused on enhancing accessibility in financial NLP. It utilizes large-scale financial web data to support real-time applications such as sentiment analysis and algorithmic trading~\cite{Lee2024fingpt}.

    \item \textbf{FinMA:} As part of the PIXIU project, FinMA is fine-tuned on FIT datasets (see section \ref{sec:fit_dataset}). It integrates structured financial knowledge into language modeling to better capture domain-specific meanings \cite{Xie2023pixiu}.

\item \textbf{InvestLM:} Developed from the LLaMA-65B base model, InvestLM is specifically adapted for investment-focused tasks, showing effectiveness in various applications including content summarization, risk evaluation processes, and financial trend analysis \cite{yang2023investlm}.

    \item \textbf{FinTral:} FinTral refers to a collection of advanced multimodal large language models constructed on the Mistral-7B architecture and fine-tuned for financial analysis. By incorporating diverse data types including text, numerical indicators, tabular information, and visual content, it is equipped to manage complex financial reasoning tasks. Empirical evaluations show that FinTral consistently outperforms ChatGPT-3.5 across all benchmarks and even exceeds GPT-4 in several cases, representing a significant leap forward in financial AI~\cite{bhatia2024fintral}.

  \item \textbf{Open-FinLLMs:} Open-FinLLMs, released by The Fin AI initiative \cite{huang2025openfinllms}, constitute a publicly available suite of financial LLMs developed to promote openness and innovation in the field. These models are based on Meta’s LLaMA 3 framework and trained on a financial corpus exceeding 52 billion tokens. The suite encompasses multiple variants, each tailored for specialized tasks:
    \begin{itemize}
        \item \textbf{FinLLaMA:} A foundational model pretrained on diverse financial data, including SEC filings, earnings calls, and market indicators. It demonstrates strong zero-shot performance in real-world financial scenarios \cite{huang2025openfinllms}.

        \item \textbf{FinLLaMA-Instruct:} An instruction-tuned version of FinLLaMA, trained on 573,000 financial instruction examples to enhance reasoning capabilities, particularly in sentiment analysis, risk assessment, and numerical reasoning \cite{huang2025openfinllms}.

        \item \textbf{FinLLaVA:} The first open-source multimodal financial LLM, capable of interpreting charts, tables, and text simultaneously, making it effective for financial decision-making and quantitative analysis \cite{huang2025openfinllms}.
    \end{itemize}

    \item \textbf{Fin-R1:} Fin-R1 is a purpose-built large language model optimized for financial reasoning and decision-making tasks. It adopts a two-phase training framework that combines supervised fine-tuning with reinforcement learning. Despite its relatively compact architecture of 7 billion parameters, Fin-R1 achieves competitive, and often superior, performance compared to larger models on financial benchmarks such as FinQA and ConvFinQA~\cite{liu2025finr1}.

\end{itemize}

These models represent the state-of-the-art in the domain of FinLLMs, each contributing unique advancements to enhance NLP applications in the financial sector. Further details on FinMA’s architecture and training are provided in Section~\ref{subsec:architecture} of Chapter~\ref{chap:methodology}.

\section{Adaptation Techniques for FinLLMs}
\label{sec:techniques}

The evolution of Financial Pre-trained Language Models (FinPLMs) and Financial Large Language Models (FinLLMs)\footnote{FinPLMs refer to smaller-scale models (typically around 110 million parameters) that are pre-trained or fine-tuned on financial data to perform domain-specific tasks. In contrast, FinLLMs are large-scale models (typically exceeding 7 billion parameters) that leverage advanced training strategies tailored for financial applications.} reflects ongoing efforts to adapt general-purpose language models to the specific demands of financial natural language processing. This adaptation is achieved through targeted pre-training and fine-tuning methodologies~\cite{Lee2024survey}. A comparative overview of these methods, as applied to four representative FinPLMs and four FinLLMs, is summarized in the work of~\citet{Lee2024survey}.

\subsection{Pre-training Strategies}

Pre-training strategies aim to endow language models with financial domain knowledge by leveraging financial or hybrid textual corpora during the initial training phase. These strategies vary in approach, depending on whether they build upon existing models or initiate training from scratch.

\textbf{Continual Pre-training.} This approach involves extending the training of a pre-existing general-domain language model by further pre-training it on financial texts, thereby incrementally adapting it to domain-specific applications. For example, FinBERT-19~\cite{Araci2019finbert} begins with BERT, originally trained on 3.3 billion general-domain tokens, and is subsequently adapted through additional training on a 29-million-word financial corpus, followed by task-specific fine-tuning for sentiment analysis. This technique enables efficient domain adaptation of general-purpose models without full re-training, making it suitable for focused financial tasks.

\textbf{Domain-Specific Pre-training from Scratch.} In this strategy, models are trained entirely on financial-domain corpora from the ground up, using standard architectures and objectives tailored to the financial context. A representative example is FinBERT-20, which is trained from scratch using a 4.9-billion-token corpus of financial communication and incorporates a dedicated vocabulary known as FinVocab \cite{Lee2024survey}. This method produces highly specialized models that capture domain-specific linguistic patterns and terminology effectively.

\textbf{Mixed-Domain Pre-training.} This method combines general-domain and financial-domain texts during pre-training to balance broad language understanding with domain relevance. For instance, FinBERT-21 utilizes both general (3.3B tokens) and financial (12B tokens) corpora and applies multi-task learning across six self-supervised objectives. Similarly, FLANG~\cite{Shah2022flang}, built on ELECTRA, adopts this approach by training on 12 billion general and 69 billion financial documents. This strategy offers a compromise between generalization and specialization, supporting a wide array of financial NLP applications.

\subsection{Fine-tuning and Adaptation Methods}

Fine-tuning and adaptation techniques are employed to tailor pre-trained language models to specific financial tasks or to increase their flexibility through prompt-based mechanisms.

\textbf{Mixed-Domain LLM with Prompt Engineering.} This method involves training large language models on a blend of general and financial corpora. Once trained, the models are used in a frozen state during inference, with prompts designed to specify tasks often expressed in natural language and optionally accompanied by examples. BloombergGPT~\cite{wu2023bloomberggpt} is an example of this approach, utilizing a BLOOM-based model pre-trained on 343 billion general tokens and 363 billion financial tokens. It was evaluated across 47 tasks, including five financial NLP tasks and 42 general NLP tasks, using prompt engineering. This method facilitates flexible deployment of models without the need for additional weight updates, thereby reducing computational overhead across varied applications.

\textbf{Instruction Fine-tuned LLM with Prompt Engineering.} This approach fine-tunes language models using instruction-based datasets that contain explicit prompts and corresponding responses. Prompt engineering is then used at inference to guide task execution. Several recent FinLLMs exemplify the diverse strategies adopted to enhance model performance in financial contexts. For instance, FinMA (PIXIU) \cite{Xie2023pixiu} fine-tunes LLaMA models with 7B and 30B parameters on the Financial Instruction Tuning (FIT) dataset (See \ref{sec:fit_dataset}), which contains 136,000 samples and targets tasks such as sentiment classification and stock movement prediction. Similarly, InvestLM \cite{yang2023investlm} leverages the LLaMA-65B architecture and undergoes specialized training on a curated dataset comprising CFA examination content and Securities and Exchange Commission (SEC) regulatory documents. In another approach, FinGPT \cite{Lee2024fingpt} applies Low-Rank Adaptation (LoRA) techniques \cite{hu2021lora} to fine-tune six open-source LLMs ~\cite{wang2023fingptinstructiontuningbenchmark} using instruction datasets tailored to financial domains. Collectively, these models demonstrate how targeted fine-tuning strategies can significantly improve model adaptability and task-specific precision, particularly for complex reasoning tasks in financial natural language processing. This method improves model adaptability and task-specific precision, making it especially effective for complex reasoning tasks in financial NLP.

\vspace{0.5em}

Table~\ref{tab:finlm_summary} provides a comparative overview of prominent financial language models. It summarizes key characteristics such as model type, parameter size, applied training techniques, data composition, and open-source accessibility adapted from \cite{Lee2024survey}.

\begin{table}[H]
\small
\centering
\begin{threeparttable}
\addtolength{\tabcolsep}{-1pt}
\begin{tabular}{p{2.3cm}|p{1.3cm}|p{1.3cm}|p{2.3cm}|p{5cm}|p{1.3cm}}
\toprule
\textbf{Model} & \textbf{Type} & \textbf{Size} & \textbf{Technique} & \textbf{Training Data} & \textbf{Open Source} \\
\midrule
FinBERT-19 & Disc & 110M & Post-PT, FT & G: 3.3B words, F: 29M words & Yes \\
FinBERT-20 & Disc & 110M & PT, FT & F: 4.9B tokens & Yes \\
FinBERT-21 & Disc & 110M & PT, FT & G: 3.3B, F: 12B words & No \\
FLANG & Disc & 110M & PT, FT & G: 3.3B, F: 696k documents & Yes \\
\midrule
BloombergGPT & Gen & 50B & PT, PE & G: 345B, F: 363B tokens & No \\
FinMA & Gen & 7B, 30B & IFT, PE & G: 1T tokens & Yes \\
InvestLM & Gen & 65B & PT, IFT, PE & G: 1.4T tokens & No\tnote{a} \\
FinGPT & Gen & 7B & IFT, PE, PEFT & G: 2T tokens & Yes \\
\bottomrule
\end{tabular}
\begin{tablenotes}
\item[a] Although the code and model weights are publicly available on GitHub, InvestLM adheres to LLaMA's license~\cite{Touvron2023llama}, which permits research-only, non-commercial use.
\end{tablenotes}
\caption{Summary of financial language models. Abbreviations: Disc = Discriminative, Gen = Generative, PT = Pretraining, FT = Fine-tuning, Post-PT = Post-pretraining, IFT = Instruction Fine-tuning, PE = Prompt Engineering, PEFT = Parameter Efficient Fine Tuning, G = General domain, F = Financial domain [Source: \cite{Lee2024survey}].}
\label{tab:finlm_summary}
\end{threeparttable}
\end{table}

\section{Financial NLP Tasks and Datasets}
\label{subsec:tasks}

Financial Large Language Models (FinLLMs) are increasingly utilized to tackle a wide range of natural language processing (NLP) challenges within the financial domain. Their applications span from sentiment analysis of market-related text to the development of automated financial advisory systems~\cite{Xie2023pixiu,Chen2024,Lee2024survey}. 

\begin{figure}[H] 
\centering
\definecolor{lightblue}{RGB}{235,245,255}
\definecolor{lightyellow}{RGB}{255,245,230}
\begin{tikzpicture}[
  font=\sffamily\footnotesize,
  box/.style={draw, rounded corners, fill=lightblue, minimum height=0.7cm, text width=8cm, align=left, drop shadow, very thick},
  boxyellow/.style={draw, rounded corners, fill=lightyellow, minimum height=0.7cm, text width=8cm, align=left, drop shadow, very thick},
  labelbox/.style={draw, rounded corners, fill=white, minimum height=0.7cm, text width=3.2cm, align=center, very thick},
  branch/.style={draw, -{Stealth[length=2mm]}, thick}
]
\node[labelbox] (sa) at (0,0) {Sentiment Analysis\\(SA)};
\node[labelbox, below=0.4cm of sa] (ie) {Information Extraction\\(IE)};
\node[labelbox, below=0.4cm of ie] (qa) {Question Answering\\(QA)};
\node[labelbox, below=0.4cm of qa] (smp) {Text-Enhanced Stock\\Movement Prediction\\(SMP)};
\node[labelbox, below=0.4cm of smp] (other) {Other Financial NLP Tasks};
\node[labelbox, below=0.4cm of other] (under) {Financial NLP Tasks\\Under-Explored for LLMs};
\node[draw, thick, fit=(sa)(ie)(qa)(smp)(other)(under), inner sep=0.3cm, label=left:{\parbox{2cm}{\centering \textbf{Financial NLP Tasks}}}] (main) {};
\node[box, right=1.4cm of sa] (sa_box) {
    Financial PhraseBank, FiQA Task 1, TweetFinSent, Finsent, SemEval-2017, StockEmotions
};
\node[box, right=1.4cm of ie] (ie_box) {
    NER FIN3, FINER-139, FinRED
};
\node[box, right=1.4cm of qa] (qa_box) {
    FiQA Task 2, FinQA, ConvFinQA, TAT-QA, DocFinQA
};
\node[box, right=1.4cm of smp] (smp_box) {
    ACL-14, ACL-18, CIKM-Stock, BigData-22, StockNet
};
\node[box, right=1.4cm of other] (other_box) {
    News Headline Classification, FOMC, FedNLP, Banking77, ECTSum, MultiLing 2019
};
\node[boxyellow, right=1.4cm of under] (under_box) {
    Financial fraud detection, compliance, robo-advisory, causality detection
};
\draw[branch] (sa.east) -- (sa_box.west);
\draw[branch] (ie.east) -- (ie_box.west);
\draw[branch] (qa.east) -- (qa_box.west);
\draw[branch] (smp.east) -- (smp_box.west);
\draw[branch] (other.east) -- (other_box.west);
\draw[branch] (under.east) -- (under_box.west);
\end{tikzpicture}
\caption{
Overview of financial NLP tasks and representative datasets for FinLLM evaluation, adapted from \citet{Chen2024}. Under-explored tasks are highlighted in yellow.
}
\label{fig:financial_nlp_tasks}
\end{figure}
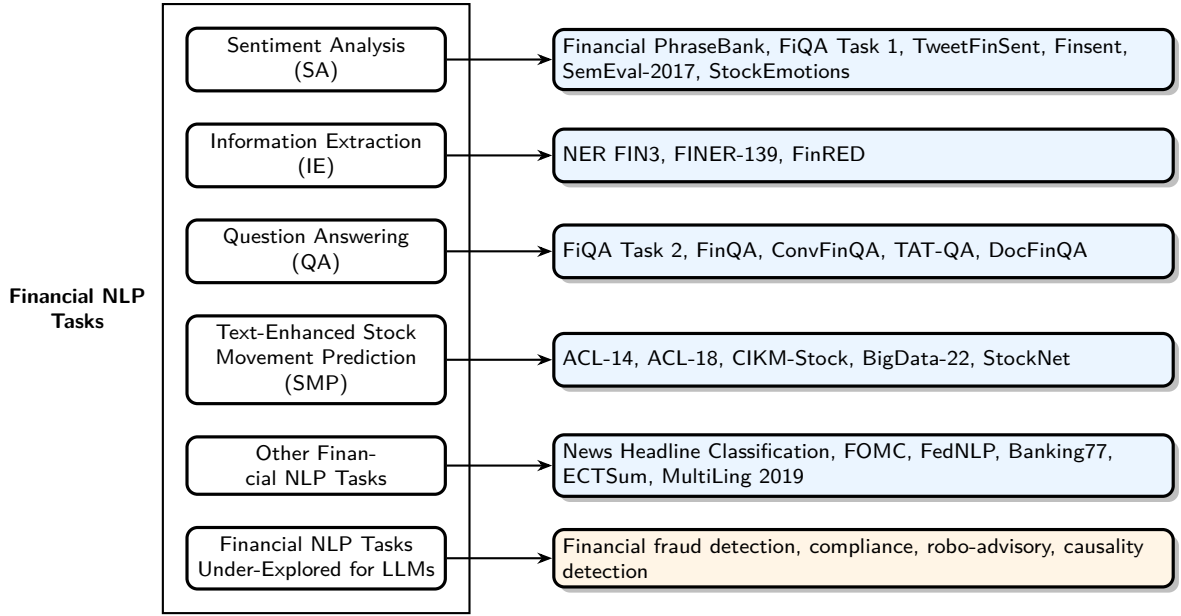

An overview of representative financial NLP tasks is depicted in Figure~\ref{fig:financial_nlp_tasks}, adapted from~\citet{Chen2024}. Table~\ref{tab:financial_nlp_tasks} complements this figure by presenting a structured summary of task categories and their associated datasets, as classified by~\citet{Lee2024survey} and aligned with widely used financial NLP benchmarks.

\begin{table}[h]
\centering
\begin{tabular}{ll}
\toprule
\textbf{Task} & \textbf{Description} \\
\midrule
Sentiment Analysis (SA) & Assessing sentiment of financial texts \\
Text Classification (TC) & Categorizing texts into predefined labels \\
Named Entity Recognition (NER) & Identifying financial entities \\
Question Answering (QA) & Providing accurate answers to financial questions \\
Stock Movement Prediction (SMP) & Predicting stock price movements \\
Text Summarization (Summ) & Generating concise summaries of financial reports \\
\bottomrule
\end{tabular}
\caption{The most common Financial NLP tasks used for evaluating FinLLMs, as mentioned by \cite{Lee2024survey}}
\label{tab:financial_nlp_tasks}
\end{table}

\subsection*{Evaluation process and scoring}
Each model is evaluated using task-specific metrics tailored to individual datasets. In sentiment analysis tasks such as FPB, models classify sentences into positive, negative, or neutral categories, with performance measured through accuracy and F1 scores against ground-truth labels. 

On Question-answering tasks like FinQA require precise numerical or textual responses to questions based on financial tables, evaluated using exact match scoring. 

For Stock movement prediction tasks, exemplified by BigData22, involve forecasting directional price changes using news sentiment and historical data, assessed through accuracy and Matthews Correlation Coefficient (MCC) to measure prediction alignment with actual market outcomes. Table \ref{tab:task_metrics} summarizes these evaluation approaches.

\begin{table}[H]
\centering
\begin{tabular}{|p{4cm}|p{4cm}|p{8cm}|}
\hline
\textbf{Task} & \textbf{Metric} & \textbf{Illustration} \\
\hline
Classification & Accuracy & Ratio of correct predictions over total predictions: (TP + TN) / Total. \\
\hline
Classification & F1 Score & Harmonic mean of precision and recall. \cite{Xie2023pixiu} report both weighted and macro versions. \\
\hline
Classification & MCC & Correlation coefficient between true and predicted classes. Ranges from -1 (inverse) to 1 (perfect). \\
\hline
Sequential Labeling & F1 Score & Computed using \texttt{seqeval}, requiring exact match on entity span and type. \cite{Xie2023pixiu} use this metric. \\
\hline
Sequential Labeling & Label F1 Score & Label-only correctness, ignores entity spans. \\
\hline
Relation Extraction & F1 Score & Harmonic mean of precision and recall on predicted relations. \\
\hline
Summarization & ROUGE-N & Measures N-gram overlap (e.g., ROUGE-1, ROUGE-2) with reference summary. \\
\hline
Summarization & ROUGE-L & Based on longest common subsequence between generated and reference summaries. \\
\hline
Question Answering & EMACC & Exact match between prediction and reference answer. \\
\hline
\end{tabular}
\caption{Predefined task metrics (adapted from \cite{Xie2023pixiu})}
\label{tab:task_metrics}
\end{table}

\section{Performance Benchmarks of FinLLMs}
\label{subsec:performance}
Financial Large Language Models (FinLLMs) are evaluated through comprehensive benchmarks like PIXIU’s FLARE \cite{Xie2023pixiu} or with the recent one FinBen \cite{xie2024finbenholisticfinancialbenchmark}, to assess their capabilities in financial NLP and prediction tasks, critical for applications in market analysis and financial advisory. As part of the literature review, this section presents the performance benchmarks from reported in literature for the six primary tasks introduced in Section~\ref{subsec:tasks}: Sentiment Analysis (SA), Text Classification (TC), Named Entity Recognition (NER), Question Answering (QA), Stock Movement Prediction (SMP), and Text Summarization (Summ). The benchmarks, is summarized in Tables~\ref{tab:performance_comparison_basic} and \ref{tab:performance_comparison_complex}. The scores presented in these tables are drawn from the literature \cite{Lee2024survey, Xie2023pixiu, yang2023investlm, wang2023fingptinstructiontuningbenchmark}, as well as from the official repositories\footnote{\url{https://github.com/adlnlp/FinLLMs?tab=readme-ov-file}} \footnote{\url{https://github.com/AI4Finance-Foundation/FinGPT}}.

\subsection{Benchmarks for Basic Financial NLP Tasks}

\cite{Lee2024survey} identifies SA, TC, and NER as basic tasks, characterized by their reliance on classification and entity extraction. Table~\ref{tab:performance_comparison_basic} presents the benchmarked performance on these tasks, as reported in the literature.

\begin{table}[htbp]
\centering
\small
\caption{Reported benchmarks for basic financial NLP tasks (SA = Sentiment Analysis, TC = Text Classification, NER = Named Entity Recognition).}
\label{tab:performance_comparison_basic}
\begin{tabular}{p{2.5cm}|p{1.8cm}|p{2cm}|p{2cm}|p{2cm}|p{2cm}}
\toprule
\multirow{2}{*}{\textbf{Model}} & \multirow{2}{*}{\textbf{Category}} & \multicolumn{2}{c|}{\textbf{SA (F1)}} & \textbf{TC (F1)} & \textbf{NER (F1)} \\
\cmidrule(lr){3-4} \cmidrule(lr){5-5} \cmidrule(lr){6-6}
 & & \textbf{FiQA-SA} & \textbf{FPB} & \textbf{Headlines} & \textbf{FIN3} \\
\midrule
BERT-base & General & -& 86\% & 97\% & 79\% \\
FinBERT-20 & Domain & -& 87\% & 97\% & 80\% \\
FLANG & Domain & -& \textbf{92\%} & \textbf{98\%} & 82\% \\
\midrule
BloombergGPT & Domain & 75\%& 51\% & 82\% & 61\% \\
\midrule
FinMA-7B-Full & Domain & 79\%& 87\% & 97\% & 69\% \\
FinMA-30B & Domain & 87\%& 88\% & \textbf{98\%} & 62\% \\
FinGPT & Domain & 87\%& 88.2\% & 94.2\% & 67.3\% \\
\midrule
ChatGPT & General & 78\%& 75\% & 91\% & 77\% \\
GPT-4 & General & 88\%& 86\% & 93\% & \textbf{83\%} \\
\bottomrule
\end{tabular}
\end{table}

For \textbf{SA}, FLANG achieving the highest Micro-F1 score of 92\% on both FiQA-SA and Financial PhraseBank (FPB), followed by FinGPT (88.2\%) and FinMA-30B (88\%). General LLMs like ChatGPT score 75\%, while BERT-base and GPT-4 reach 86\%.

In \textbf{TC}, FLANG and FinMA-30B attain 98\% F1 on the Headlines dataset, with BERT-base, FinBERT-20 and FinMA-7B-Full at 97\%, and GPT-4 at 93\%. For \textbf{NER}, GPT-4 leads with 83\% F1 on the FIN3 dataset, followed by FLANG (82\%), while FinMA-30B and BloombergGPT score 62\% and 61\%.

\subsection{Benchmarks for complex Financial NLP Tasks}

Same authors \cite{Lee2024survey} classifies QA, SMP, and Summ as complex tasks, requiring numerical reasoning and generative capabilities. Table~\ref{tab:performance_comparison_complex} summarizes the reported benchmarks for these tasks.

\begin{table}[htbp]
\centering
\small
\begin{tabular}{p{2.6cm}|p{1.5cm}|p{1.5cm}|p{1.9cm}|p{5cm}|p{1.5cm}}
\toprule
\multirow{2}{*}{\textbf{Model}} & \multirow{2}{*}{\textbf{Category}} & \multicolumn{2}{c|}{\textbf{QA (EM)}} & \textbf{SMP (Acc)} & \textbf{Summ (R1)} \\
\cmidrule(lr){3-4} \cmidrule(lr){5-5} \cmidrule(lr){6-6}
 & & \textbf{FinQA} & \textbf{ConvFinQA} & \textbf{BigData22/ACL18/CIKM18} & \textbf{ECTSum} \\
\midrule
BERT-base & General & -- & -- & -- & -- \\
FinBERT-20 & Domain & -- & -- & -- & -- \\
FLANG & Domain & -- & -- & -- & -- \\
\midrule
BloombergGPT & Domain & -& 43\% & -- & -- \\
\midrule
FinMA-7B-Full & Domain & 4\%& 20\% & 53\%  /  56\%  /  53\%   & 8\% \\
FinMA-30B & Domain & 11\%& 40\% & 47\%  /  49\%  /  43\%   & -- \\
FinGPT & Domain & -- & -- & -- & -- \\
\midrule
ChatGPT & General & 49\%& 60\% & 53\%  /  50\%  /  55\%   & 21\% \\
GPT-4 & General & \textbf{76\%} & \textbf{76\%} & \textbf{54\%  /  52\%  /  57\%} & 30\% \\
\midrule
SOTA (Task-specific) & General & -- & -- & 55\%  /  61\%  /  59\% & \textbf{47\%} \\
Human Expert& Human& 91\%& 89\% & -- & -- \\
Gene& Human& 51\% & 47\% & -- & -- \\
\bottomrule
\end{tabular}
\caption{Reported benchmarks for advanced financial NLP tasks (QA = Question Answering, SMP = Stock Movement Prediction, Summ = Text Summarization).}
\label{tab:performance_comparison_complex}
\end{table}

In \textbf{QA}, GPT-4 achieving 76\% EM on FinQA and ConvFinQA, with task-specific SOTA models at 89\%. FinMA-30B scores 40\%, BloombergGPT 43\%, and ChatGPT 60\%. For \textbf{SMP}, GPT-4 attains 54\% accuracy on BigData22, FinMA-7B 52\%, and SOTA models 58\%, with FinMA-30B at 46\%. In \textbf{Summ}, SOTA models lead with 47\% ROUGE-1 on ECTSum, followed by GPT-4 (30\%) and ChatGPT (21\%), while FinMA-7B scores 8\%.

\subsection{FinMA in the Literature}

FinMA has also been widely studied in the literature \cite{Lee2024survey, Xie2023pixiu}. Benchmarked using the FLARE framework, it demonstrates competitive performance in sentiment analysis (SA) and text classification (TC), often achieving results comparable to GPT-3.5 on stock sentiment tasks \cite{Xie2023pixiu}. Further details regarding its training and evaluation procedures will be detailled  in~\ref{chap:methodology}. Moreover, FinMA has been explored in advanced tasks such as relation extraction (e.g., the FinRED dataset) and multimodal understanding (e.g., the MAEC dataset), which, while less frequently benchmarked, represent promising directions for future research \cite{Lee2024survey, Chen2024}.

\chapter{Methodology}
\label{chap:methodology}

This chapter presents the methodology used to develop and evaluate \textbf{FinMA}, a financial large language model built within the PIXIU framework \cite{Xie2023pixiu}. It covers the model architecture, the Financial Instruction Tuning (FIT) dataset, the fine-tuning process, and the Financial Language Understanding and Prediction Evaluation Benchmark (FLARE). The evaluation strategy, analysis techniques, and reproducibility considerations are also described to ensure transparency and replicability.

\section{FinMA Architecture}
\label{subsec:architecture}

FinMA is built upon Meta’s LLaMA architecture, a transformer-based, decoder-only large language model optimized for efficiency and scalability \cite{Touvron2023llama}. Two backbone models are used: LLaMA-7B and LLaMA-30B, which differ in parameter count and depth.

\begin{itemize}
    \item \textbf{Transformer Layers:} 32 layers for LLaMA-7B and 60 layers for LLaMA-30B, each composed of multi-head self-attention and feed-forward networks using SwiGLU activation.
    \item \textbf{Embedding Size:} 4,096 for LLaMA-7B and 6,656 for LLaMA-30B, supporting rich contextual representation.
    \item \textbf{Attention Mechanism:} Multi-head self-attention with Root Mean Square Layer Normalization (RMSNorm), no dropout, and enhanced numerical stability.
    \item \textbf{Positional Encoding:} Rotary Positional Embeddings (RoPE), enabling context lengths up to 2,048 tokens.
\end{itemize}

FinMA includes three model variants \cite{Xie2023pixiu}, all fine-tuned on the Financial Instruction Tuning (FIT)(See \ref{sec:fit_dataset})  dataset:

\begin{itemize}
    \item \textbf{FinMA-7B:} Fine-tuned from LLaMA-7B using instruction tuning on financial NLP tasks, such as in sentiment analysis, headline classification, named entity recognition (NER), and question answering.
    \item \textbf{FinMA-30B:} Fine-tuned from LLaMA-30B using instruction tuning on the same financial NLP tasks, offering enhanced capacity and performance.
    \item \textbf{FinMA-7B-full:} Fine-tuned from LLaMA-7B using full instruction tuning data, incorporating financial natural language understanding (NLP) and prediction applications (e.g., anticipating stock market fluctuations).
\end{itemize}

All FinMA variants were optimized using the AdamW algorithm, configured with an initial learning rate of 8e$^{-6}$, a weight decay of 1e$^{-5}$, and a warm-up schedule covering 5\% of the total training steps. The maximum sequence length was capped at 2,048 tokens to accommodate long-form financial documents \cite{Xie2023pixiu}.

The FinMA-7B model was trained over 15 epochs using 8 NVIDIA A100 GPUs with 40GB of memory. A variant version, FinMA-7B-full, underwent a shorter fine-tuning regime of three epochs under identical hardware conditions. In contrast, FinMA-30B required a more extensive training setup completing 20 epochs with a reduced batch size of 24 utilizing distributed training across 128 A100 40GB GPUs \cite{Xie2023pixiu}. Table \ref{tab:finma_variants} provides an overview of these configurations across the variants of FinMA.


\begin{table}[ht]
\centering
\caption{Architectural and training configurations of FinMA variants \cite{Xie2023pixiu, Touvron2023llama}.}
\label{tab:finma_variants}
\begin{tabular}{l|c|c|c}
\toprule
\textbf{Feature} & \textbf{FinMA-7B} & \textbf{FinMA-7B-full} & \textbf{FinMA-30B} \\
\midrule
Base Model & LLaMA-7B & LLaMA-7B & LLaMA-30B \\
Parameters & $\sim$7B & $\sim$7B & $\sim$30B \\
Transformer Layers & 32 & 32 & 60 \\
Attention Heads & 32 & 32 & 52 \\
Hidden Size & 4096 & 4096 & 6656 \\
Adaptation & Full fine-tuning & Full fine-tuning & Full fine-tuning \\
Epochs & 15 & 3 & 20 \\
Batch Size & 32 & 32 & 24 \\
GPUs Used & 8 $\times$ A100 40GB & 8 $\times$ A100 40GB & 128 $\times$ A100 40GB \\
Input Length & 2048 tokens & 2048 tokens & 2048 tokens \\
Tasks & NLP & NLP + Prediction & NLP \\
\bottomrule
\end{tabular}
\end{table}

By leveraging the LLaMA architecture and applying comprehensive instruction tuning, FinMA effectively adapts large language models to the financial domain, demonstrating robust performance across diverse financial natural language processing and prediction tasks.

\section{Financial Instruction Tuning (FIT) Dataset}
\label{sec:fit_dataset}

The Financial Instruction Tuning (FIT) dataset was specifically designed by the creators of FinMA \cite{Xie2023pixiu} to address the scarcity of high-quality, instruction-tuning data tailored to the financial domain. It is among the first datasets to target this need, thereby enabling the training of models capable of responding to financial queries using diverse instruction formats.

The dataset leverages publicly accessible financial data and integrates multiple task types and modalities, aiming to support a broad spectrum of real-world applications. FIT contains 136,609 instruction samples that were drawn from nine publicly available sources. These samples span five task categories: sentiment analysis (SA), news headline classification (NC), named entity recognition (NER), question answering (QA), and stock movement prediction (SMP). The data sources include news headlines, tweets, earnings reports, and regulatory filings such as those from the SEC. Moreover, FIT incorporates a multimodal structure, with instances comprising textual content, tables, and time series data nabling the model to develop more nuanced reasoning abilities across different financial contexts.

Table~\ref{tab:fit} summarizes the datasets, task types, number of samples, and associated modalities.



\subsection{FIT construction}

\begin{table}[ht]
\centering
\caption{Datasets in the Financial Instruction Tuning (FIT) corpus.}
\label{tab:fit}
\begin{tabular}{l l r r l}
\toprule
\textbf{Dataset} & \textbf{Task} & \textbf{Raw} & \textbf{Instruction} & \textbf{Modalities} \\
\midrule
Financial Phrase Bank & Sentiment Analysis         & 4,845   & 48,450  & Text \\
FiQA-SA                      & Sentiment Analysis         & 1,173   & 11,730  & Text \\
Gold News Headlines          & News Headline Classification & 11,412 & 11,412  & Text \\
FIN Agreements               & Named Entity Recognition   & 1,366   & 13,660  & Text \\
FinQA                        & Question Answering         & 8,281   & 8,281   & Text, Tables \\
ConvFinQA                    & Question Answering         & 3,892   & 3,892   & Text, Tables \\
BigData22                    & Stock Movement Prediction  & 7,164   & 7,164   & Text, Time-Series \\
ACL18                        & Stock Movement Prediction  & 27,053  & 27,053  & Text, Time-Series \\
CIKM18                       & Stock Movement Prediction  & 4,967   & 4,967   & Text, Time-Series \\
\bottomrule
\end{tabular}
\end{table}

\subsection{Dataset Format}

Each FIT instance follows a structured JSON format suitable for instruction tuning:

\begin{lstlisting}[language=json]
{
    "id": "unique_id",
    "conversations": [
        {
            "from": "human",
            "value": "Instructional prompt + input text"
        },
        {
            "from": "agent",
            "value": "Expected response"
        }
    ],
    "text": "Raw input text or table",
    "label": "Ground truth label"
}
\end{lstlisting}

\begin{itemize}
    \item \texttt{"id"}: Unique identifier
    \item \texttt{"conversations"}: Human-agent instruction turns
    \item \texttt{"text"}: Input data to be analyzed
    \item \texttt{"label"}: Expected classification or response
\end{itemize}

\subsection{Prompt Engineering and Domain Expertise}

To enhance instruction-following capabilities, FIT reformulates raw data into instruction-based triplets: \texttt{(instruction, input, response)}. Each task includes a diverse set of carefully crafted prompts, designed by \textit{financial domain experts}, ensuring realism and generalization \cite{Xie2023pixiu}.


\begin{framed}
\begin{minipage}{\linewidth}
\textbf{Raw Data Example (from FPB):}
\begin{quote}
``The company reported a 15\% increase in quarterly profits.'' $\rightarrow$ Positive
\end{quote}

This raw example is converted into an instruction-based format for training, as shown in Listing~\ref{lst:fit_sample} below.

\begin{lstlisting}[caption={Sample instruction-tuning entry from FIT (FPB)},label={lst:fit_sample}]
{
  "instruction": "Determine the sentiment expressed in the following financial news excerpt:",
  "text": "The company reported a 15% increase in quarterly profits.",
  "response": "Positive"
}
\end{lstlisting}
\end{minipage}
\end{framed}

This design helps the model learn how to follow instructions, which is important for real-world usage. For example, instead of only seeing direct labels like ``positive'', the model is trained to understand different ways people may ask for sentiment analysis, such as:
\begin{lstlisting}[language=json,frame=single,caption={Examples of sentiment analysis prompts},label={lst:sentiment_examples}]
    "Classify the sentiment of this statement:",
    "What is the sentiment expressed in the following sentence?",
    "Is the tone positive, negative, or neutral?",
    etc.
\end{lstlisting}

As can be referred to in Table~\ref{tab:prompts} for the rest of the task.




\begin{table}[H]
\centering
\caption{Illustrative prompts corresponding to selected financial NLP datasets. In FiQA-SA, \texttt{\{category\}} is a placeholder for content such as news headlines or tweets. In BigData22, \texttt{\{tid\}} and \texttt{\{point\}} represent stock tickers and time points, respectively~\cite{Xie2023pixiu}.}
\label{tab:prompts}
\begin{tabular}{l p{14cm}}
\toprule
\textbf{Dataset} & \textbf{Example Prompt} \\
\midrule
FPB & Determine the sentiment conveyed in the financial news statement: negative, positive, or neutral. For instance, ``Stocks plummeted after the scandal'' should be labeled as negative. \\
\midrule

FiQA-SA & Assess the sentiment of the financial \texttt{\{category\}} (e.g., headline or tweet) and classify it as Positive, Negative, or Neutral. \\
\midrule

Headline & Does the given headline refer to gold price movements? Respond with ``Yes'' or ``No''. \\
\midrule

NER & Extract named entities from U.S. SEC filings and label them as Person (PER), Organization (ORG), or Location (LOC). Use the format: ``entity name, entity type''. Example: ``Elon Musk, PER; SpaceX, ORG; Cape Canaveral, LOC''. \\
\midrule

FinQA & Provide a concise answer to the financial question using data and reasoning from available sources. \\
\midrule

ConvFinQA & Use the given pretext, table, and document to answer the final financial question, applying necessary calculations and contextual understanding. \\
\midrule

BigData22 & Based on market signals and social media activity, predict whether the stock \texttt{\{tid\}} will increase or decrease at time \texttt{\{point\}}. Answer with: Rise or Fall. \\
\bottomrule
\end{tabular}
\end{table}

\section{FLARE: Financial Evaluation Benchmark}
\label{sec:flare_benchmark}

The \textbf{FLARE benchmark} (Financial Language Understanding and Prediction Evaluation) is also a benchmark specifically designed by the authors of FinMA \cite{Xie2023pixiu} to evaluate the performance of large language models (LLMs) on both financial natural language understanding and financial prediction tasks. Unlike earlier benchmarks such as FLUE (shah2022flue), which focus exclusively on NLP tasks, FLARE integrates financial prediction tasks such as stock movement forecasting, providing a more comprehensive assessment of LLM capabilities in real-world financial applications.

FLARE is constructed from the FIT dataset (see Section~\ref{sec:fit_dataset}). The authors \cite{Xie2023pixiu} follow a principled approach by randomly selecting validation and test subsets from FIT, in line with established evaluation protocols \cite{guo2023evaluating}. This sampling strategy not only ensures statistical soundness but also facilitates meaningful comparison with proprietary models. In particular, the number and distribution of test samples in FLARE are aligned with those used in the evaluation of BloombergGPT \cite{wu2023bloomberggpt}, whose original test data is not publicly released. This alignment allows researchers to benchmark open-source models under comparable settings.

The benchmark covers five task categories as FIT[\ref{sec:fit_dataset}]: sentiment analysis, news headline classification, named entity recognition (NER), question answering (QA), and stock movement prediction (SMP). Each task is associated with one or more datasets and evaluated using appropriate metrics[\ref{tab:task_metrics}]. Sentiment analysis tasks (FPB, FiQA-SA) are evaluated using accuracy and F1 score; QA tasks (FinQA, ConvFinQA) use exact match (EM) accuracy; stock prediction tasks (BigData22, ACL18, CIKM18) are evaluated using both accuracy and Matthews correlation coefficient (MCC). Table~\ref{tab:flare} summarizes the tasks, datasets, and evaluation metrics.


\begin{table}[ht]
\centering
\caption{FLARE benchmark tasks, datasets, and evaluation metrics.}
\label{tab:flare}
\begin{tabular}{l l l}
\toprule
\textbf{Task} & \textbf{Dataset(s)} & \textbf{Evaluation Metric(s)} \\
\midrule
Sentiment Analysis  & FPB, FiQA-SA            & F1 score, Accuracy \\
News Classification & Gold News Headlines     & Average F1 score \\
Named Entity Recognition & FIN Agreements      & Entity-level F1 score \\
Question Answering  & FinQA, ConvFinQA        & Exact Match (EM) \\
Stock Movement Prediction & BigData22, ACL18, CIKM18 & Accuracy, MCC \\
\bottomrule
\end{tabular}
\end{table}



\subsection{Evaluation Protocol}
\label{subsec:evaluation_protocol}

The evaluation follows established practices in financial NLP, \cite{guo2023evaluating, wu2023bloomberggpt}:

\textbf{Zero-shot vs Few-shot Settings:}
\begin{itemize}
    \item \textbf{Zero-shot:} Model receives only the instruction and input, with no examples shown before prediction. All results for FinMA are reported in the zero-shot setting \cite{Xie2023pixiu}.
    \item \textbf{Few-shot:} Model is given a few examples of the task before making its prediction.
    \begin{itemize}
        \item BloombergGPT: \textbf{20-shot} on FIN dataset, \textbf{5-shot} on FPB and FiQA-SA
        \item Other baselines: \textbf{5-shot} on News dataset
    \end{itemize}
\end{itemize}

Some baseline models that were not instruction-tuned had trouble generating answers in the expected format (e.g., they produce long text when only a label like "positive" is expected). In such cases, results were evaluated by humans instead of automatically \cite{Xie2023pixiu}. This ensures a fair comparison between FinMA and BloombergGPT on the FLARE benchmark following \cite{Xie2023pixiu}.


\section{Fine-Tuning Procedure}
\label{subsec:fine_tuning}

FinMA models are fine-tuned using instruction-based learning on the FIT dataset (see Section~\ref{sec:fit_dataset}). The goal is to align the model with financial domain tasks by training it to follow natural language instructions across multiple modalities and task types.

As seen in \ref{sec:fit_dataset}, each training sample follows a standard format:
\begin{quote}
    \texttt{(instruction, input, response)}
\end{quote}
This structure teaches the model to produce context-aware outputs in response to instructions, preparing it for zero-shot generalization. And this format enables FinMA to align language understanding with financial task objectives in a natural and structured way.

\begin{figure}[ht]
    \centering
\includegraphics[width=0.9\textwidth]{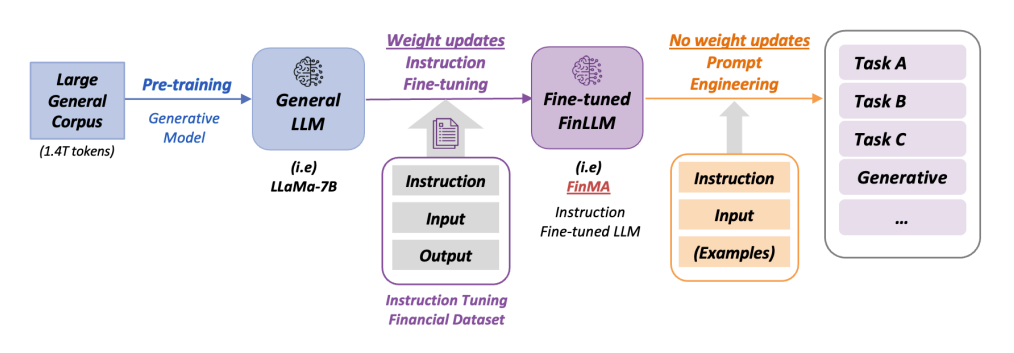}
    \caption{Instruction tuning pipeline for financial datasets \cite{Lee2024survey}.}
    \label{fig:instruction_tuning}
\end{figure}

\vspace{1cm}

\subsection*{Training Infrastructure}

FinMA uses a causal language modeling objective, where the model predicts tokens autoregressively. Training is performed using the AdamW optimizer, with \textbf{Learning rate:} $8 \times 10^{-6}$
    and for \textbf{Weight decay:} $1 \times 10^{-5}$ as mentioned in \ref{subsec:architecture} and
training was conducted using 8 NVIDIA A100 GPUs (40GB each), with implementation based on PyTorch and Hugging Face Transformers.

\begin{table}[H]
\centering
\begin{tabular}{l c c l}
\toprule
\textbf{Model Variant} & \textbf{Hardware} & \textbf{Training Duration} & \textbf{Environment} \\
\midrule
FinMA-7B & 8 $\times$ A100 40GB & 15 epochs & Google Colab Pro \\
FinMA-7B-full & 8 $\times$ A100 40GB & 3 epochs & Google Colab Pro \\
FinMA-30B & 128 $\times$ A100 40GB & 20 epochs & Distributed Training \\
\bottomrule
\end{tabular}
\caption{Training infrastructure specifications for FinMA variants.}
\label{tab:training_infrastructure}
\end{table}
    
All variants are fully fine-tuned (no adapters or LoRA\cite{hu2021lora} techniques are used) ensuring deep adaptation to financial reasoning.

\subsection*{Prompt diversity and instruction design}

As described in Section~\ref{sec:fit_dataset}, domain experts created multiple instruction templates per task to improve generalization. For smaller datasets, up to 10 instructions were applied to each raw data point; for larger datasets, one instruction was randomly sampled to maintain scalability.

This diversity ensures that the model learns to handle different phrasings of the same task. For example, in sentiment analysis, prompts may vary from "What is the sentiment?" to "Classify the tone of this statement." This strategy contributes directly to FinMA’s strong zero-shot performance (see Section~\ref{sec:flare_benchmark}).

\section{Inference and Evaluation Framework}
\label{sec:eval_framework}

In this section, we describe the evaluation process for inference and our setup for inferring the model across tasks to reproduce the results and highlight our contributions. This also serves as a basis for the discussion of results presented in Chapter \ref{chap:results_analysis}.

\subsection{Environment Setup}

To ensure reproducibility and avoid package conflicts, a dedicated Python virtual environment was created and activated. All dependencies were installed within this environment.

\begin{lstlisting}[language=bash]
!pip install virtualenv
!virtualenv finma_venv
\end{lstlisting}

\subsection{Downloading and Preparing Tools}

Two main repositories were cloned:
\begin{itemize}
    \item \textbf{BARTScore\footnote{\url{https://github.com/neulab/BARTScore}}}: for advanced text generation evaluation.
    \item \textbf{PIXIU}: contains the financial language models and evaluation scripts.
\end{itemize}

\begin{lstlisting}[language=bash]
!git clone https://github.com/neulab/BARTScore.git --recursive
!git clone https://github.com/The-FinAI/PIXIU.git --recursive
\end{lstlisting}

\subsection{Model and Metric Preparation}

The FinMA-7B-full model\footnote{\url{https://huggingface.co/ChanceFocus/finma-7b-full}} was downloaded from Hugging Face, and BARTScore was set up with its pretrained checkpoint.

\begin{lstlisting}[language=Python]
from huggingface_hub import login
login(token="your_hf_token")

from bart_score import BARTScorer
bart_scorer = BARTScorer(device='cuda:0' if torch.cuda.is_available() else 'cpu',
                         checkpoint='facebook/bart-large-cnn')
score = bart_scorer.score(['This is interesting.'], ['This is fun.'], batch_size=4)
print("BARTScore:", score)
\end{lstlisting}

\subsection{Integration of Evaluation Tools}

To ensure BARTScore works with the evaluation scripts, the \texttt{eval.py} file was edited to include the correct path for the BARTScore module if not already present.

\begin{lstlisting}[language=Python]
file_path = "/kaggle/working/PIXIU/src/eval.py"
path_to_add = "/kaggle/working/PIXIU/src/metrics/BARTScore"
with open(file_path, "r") as f:
    content = f.read()
if f"sys.path.append('{path_to_add}')" not in content:
    new_content = f"import sys\nsys.path.append('{path_to_add}')\n" + content
    with open(file_path, "w") as f:
        f.write(new_content)
\end{lstlisting}

\subsection{Running Financial NLP and Prediction Tasks}

The FinMA-7B-full model was evaluated across nine datasets covering six financial NLP tasks using the PIXIU benchmark. Inference was performed using the following command:

\begin{lstlisting}[language=bash]
!python eval.py \
  --model "hf-causal-llama" \
  --model_args "use_accelerate=True,pretrained=TheFinAI/finma-7b-full,tokenizer=TheFinAI/finma-7b-full,use_fast=False" \
  --tasks "flare_fiqasa,flare_fpb,flare_ner,flare_headline,flare_finqa,flare_convfinqa,flare_bigdata22,flare_skm,flare_cikm,flare_ectsumm"
\end{lstlisting}

\subsection{Special Handling for ConvFinQA Evaluation}
\label{subsec:convfinqa_evaluation}

For the \textbf{ConvFinQA} dataset, we implemented a dedicated inference pipeline. The default evaluation script (\texttt{eval.py}) provided in the PIXIU repository could not be executed as intended due to technical issues previously reported by other users\footnote{\url{https://github.com/chancefocus/PIXIU/issues}}. As such, a custom evaluation procedure was implemented for assessing the zero-shot performance of the FinMA-7B-full model on this benchmark. The detailed process is outlined below.

\textbf{a. Model Initialization}. The FinMA-7B-full model was loaded from Hugging Face using the Transformers library. To support efficient inference, the model was cast to \texttt{float16} precision and deployed with automatic device mapping. A generation pipeline was configured, and \texttt{pad\_token\_id} was explicitly defined to prevent runtime errors during generation.

\textbf{b. Dataset Loading.} We loaded the \texttt{flare-convfinqa} test split using the \texttt{datasets} library. Each instance included a natural language financial question (\texttt{query}), a numerical answer (\texttt{answer}), the dialogue turn number (\texttt{turn}), and a unique identifier for the multi-turn dialogue (\texttt{dialogue\_id}).

\textbf{c. Prompt Design}. Each question was embedded in a structured prompt as follows:

\begin{quote}
\texttt{Question: \{query\}}

\texttt{Context: \{context\}}

\texttt{Answer with only a number:}

\texttt{Answer:}
\end{quote}

In most cases, the context field was left empty, as the questions are self-contained.

\textbf{d. Answer Generation}. The prompts were tokenized and passed to the FinMA-7B-full model. Responses were generated with a maximum of 32 new tokens. The output was then parsed to extract only the predicted value following the final \texttt{Answer:} marker.

\textbf{e. Evaluation Metrics}. To quantitatively assess performance, we computed two metrics widely used in question answering: \textbf{Exact Match (EM)}
and \textbf{F1 Score} as summurize in Table \ref{tab:task_metrics}. All answers were normalized by converting to lowercase, removing punctuation, and collapsing multiple whitespaces.

\subsection{Metrics and Evaluation Protocol}
\label{subsec:metrics_evaluation}

The evaluation relied on predefined metrics per task, as summarized in \ref{tab:task_metrics} (Section \ref{sec:fit_dataset}). These include F1 score, Accuracy, Exact Match (EM) and ROUGE1. Readers are referred to that section for definitions of those metrics.

\subsection{Experimental Environment}
\label{subsec:experimental_environment}

\begin{table}[h]
\centering
\caption{Kaggle-based experimental environment specifications.}
\label{tab:experimental_environment}
\begin{tabular}{p{2.5cm} p{4cm} p{5cm}}
\toprule
\textbf{Component} & \textbf{Specification} & \textbf{Purpose} \\
\midrule
Hardware &  T4 GPUs (16GB), up to 4 CPU cores & Model inference in Kaggle kernels \\
Software & PyTorch, HuggingFace Transformers, Python 3.11 & Deep learning and model management \\
Platform & Kaggle Notebooks (free tier) & Computational environment with GPU/TPU support \\
Storage & 20GB temporary disk, 5GB persistent output & Dataset and model storage \\
\bottomrule
\end{tabular}
\end{table}

\section{Reproducibility and Open Science}
\label{sec:reproducibility}

The PIXIU repository\footnote{\url{https://github.com/chancefocus/PIXIU}}  provides comprehensive reproducibility support \cite{Xie2023pixiu}. Additional resources, including custom scripts and documentation, are available in our GitHub repository\footnote{\url{https://github.com/AbdelkaderYS}}.

\chapter{Results and Analysis}
\label{chap:results_analysis}

This chapter evaluates the performance of \textbf{FinMA-7B-full} on the FLARE benchmark using the PIXIU framework \cite{Xie2023pixiu}, as outlined in Chapter~\ref{chap:methodology} (Section~\ref{sec:eval_framework}). The evaluation covers \textbf{sentiment analysis (SA)}, \textbf{news headline classification (TC)}, \textbf{named entity recognition (NER)}, \textbf{question answering (QA)}, \textbf{stock movement prediction (SMP)}, and \textbf{text summarization (Summ)} in zero-shot, 5-shot, and 20-shot settings, following the evaluation methodology of BloombergGPT \cite{wu2023bloomberggpt}. Results are compared against baselines from \citet{Xie2023pixiu} and other models, including BloombergGPT, GPT-4, and ChatGPT, as reviewed in Chapter~\ref{chap:literature_review} (Tables~\ref{tab:performance_comparison_basic}, \ref{tab:performance_comparison_complex}).

\section{Presentation of Results}
\label{sec:results}

\subsection{Baseline Definition and Re-evaluation}
\label{subsec:baseline}

In this study, the \textit{baseline} refers to the zero-shot performance of FinMA-7B-full as reported in \citet{Xie2023pixiu} within the FLARE benchmark. This baseline measures the model's inherent capabilities without task-specific examples, a standard approach for assessing the generalizability of large language models (LLMs) in financial NLP \cite{brown2020language, Lee2024survey}. The re-evaluation of FinMA-7B-full in zero-shot settings verifies the reproducibility of these results, ensures alignment with the PIXIU framework, and extends the analysis to 5-shot and 20-shot settings for comprehensive benchmarking \cite{guo2023evaluating}. This methodology confirms the model’s robustness, aligns with industry standards such as BloombergGPT \cite{wu2023bloomberggpt}, and provides new insights into FinMA’s performance across diverse financial tasks.

Performance results are summarized in Tables~\ref{tab:financial_nlp_results_1} (SA, TC, NER) and \ref{tab:financial_nlp_results_2} (QA, SMP, Summ), presenting zero-shot, 5-shot, and 20-shot performance metrics compared to the zero-shot baseline from \citet{Xie2023pixiu}.

\begin{longtable}{@{}p{4cm}p{2cm}p{2cm}p{2cm}p{2cm}p{3cm}@{}}
\caption{Financial NLP Benchmarking Results: Sentiment Analysis, Classification, and NER}\label{tab:financial_nlp_results_1}\\
\toprule
\multirow{2}{*}{\textbf{Task \& Dataset}} & \multicolumn{4}{c}{\textbf{Performance Scores (\%)}} & \multirow{2}{*}{\textbf{Best vs Baseline}} \\
\cmidrule(lr){2-5}
& \textbf{Baseline (0-Shot)} & \textbf{Zero-Shot} & \textbf{5-Shot} & \textbf{20-Shot} & \\
\midrule
\endfirsthead
\multicolumn{6}{c}{\textbf{Table \thetable{} (continued)}} \\
\toprule
\multirow{2}{*}{\textbf{Task \& Dataset}} & \multicolumn{4}{c}{\textbf{Performance Scores}} & \multirow{2}{*}{\textbf{Best vs Baseline}} \\
\cmidrule(lr){2-5}
& \textbf{Baseline (0-Shot)} & \textbf{0-Shot} & \textbf{5-Shot} & \textbf{20-Shot} & \\
\midrule
\endhead
\midrule
\multicolumn{6}{r}{\textit{Continued on next page}} \\
\endfoot
\bottomrule
\endlastfoot
\multicolumn{6}{@{}l@{}}{\cellcolor{gray!10}\textbf{\large SENTIMENT ANALYSIS}} \\[0.5ex]
\midrule
\textbf{FiQA-SA} & & & & & \\
\quad $\circ$ F1-Score & 79.0 & \textcolor{red}{\textbf{83.5}} & 82.6 & 81.2 & \textcolor{green}{+4.5} \\
\addlinespace[0.3ex]
\textbf{FPB} & & & & & \\
\quad $\circ$ F1-Score & 87.0 & \textcolor{red}{\textbf{93.9}} & 93.4 & 93.4 & \textcolor{green}{+6.9} \\
\midrule
\multicolumn{6}{@{}l@{}}{\cellcolor{gray!10}\textbf{\large NEWS HEADLINE CLASSIFICATION}} \\[0.5ex]
\midrule
\textbf{Headlines} & & & & & \\
\quad $\circ$ Avg F1-Score & 97.0 & \textcolor{red}{\textbf{97.5}} & 93.5 & -- & \textcolor{green}{+0.5} \\
\midrule
\multicolumn{6}{@{}l@{}}{\cellcolor{gray!10}\textbf{\large NAMED ENTITY RECOGNITION}} \\[0.5ex]
\midrule
\textbf{Financial NER} & & & & & \\
\quad $\circ$ Entity F1-Score & 69.0 & 53.9 & \textcolor{red}{\textbf{64.1}} & 65.8 & \textcolor{orange}{-4.9} \\
\bottomrule
\end{longtable}

\begin{figure}[H]
    \centering
    \includegraphics[width=1\linewidth]{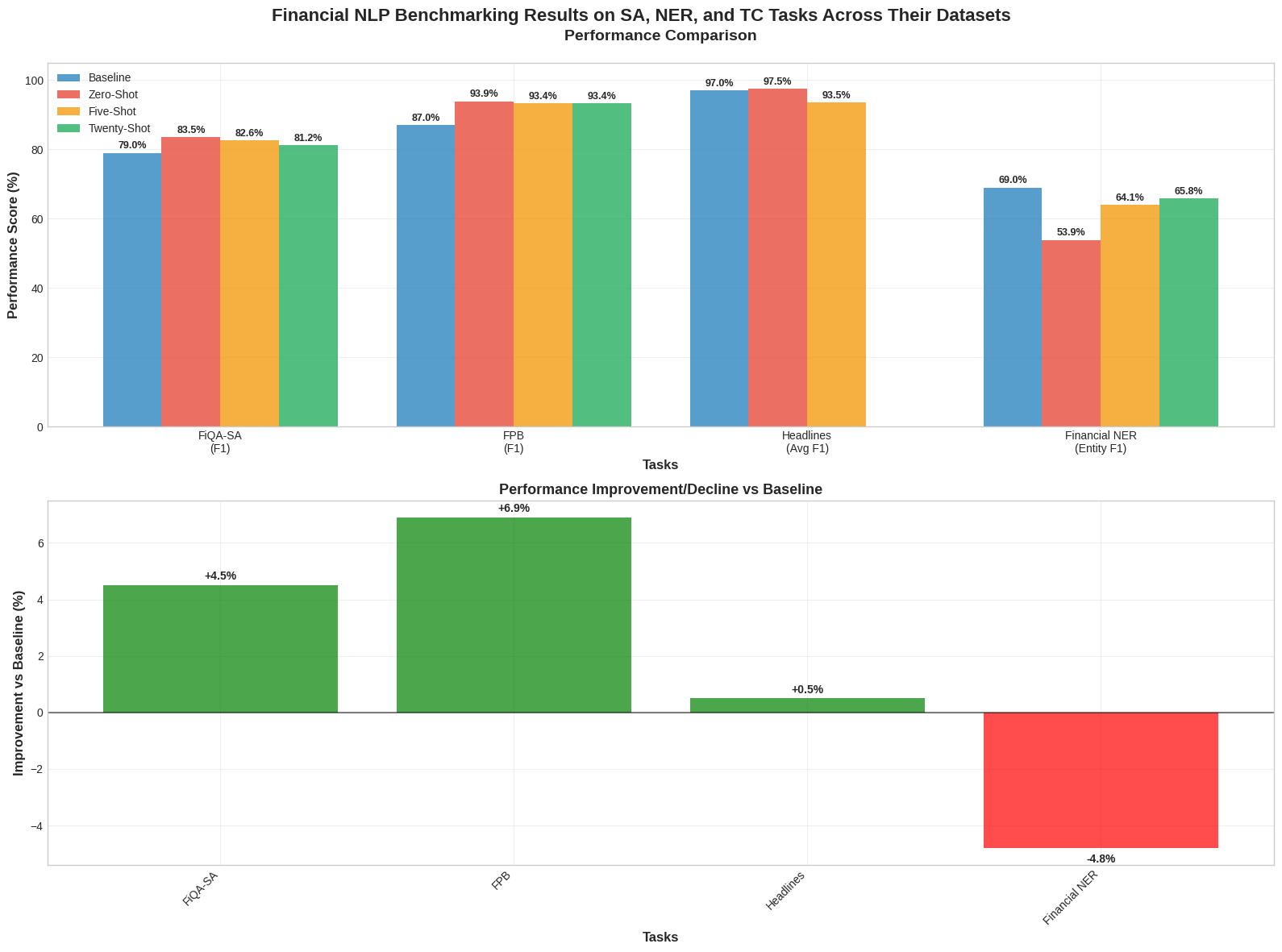}
    \caption{Zero-shot and Few-shot F1 Scores for Sentiment Analysis (FiQA-SA, FPB), News Headline Classification (Headlines), and Named Entity Recognition (Financial NER).}
    \label{fig:sa_tc_ner_results}
\end{figure}

\vspace{-0.1cm}

\begin{longtable}[H]{@{}p{4cm}p{2cm}p{2cm}p{2cm}p{2cm}p{3cm}@{}}
\caption{Financial NLP Benchmarking Results: Question Answering, Stock Prediction, and Summarization}\label{tab:financial_nlp_results_2}\\
\toprule
\multirow{2}{*}{\textbf{Task \& Dataset}} & \multicolumn{4}{c}{\textbf{Performance Scores (\%)}} & \multirow{2}{*}{\textbf{Best vs Baseline}} \\
\cmidrule(lr){2-5}
& \textbf{Baseline (0-Shot)} & \textbf{Zero-Shot} & \textbf{5-Shot} & \textbf{20-Shot} & \\
\midrule
\endfirsthead
\multicolumn{6}{c}{\textbf{Table \thetable{} (continued)}} \\
\toprule
\multirow{2}{*}{\textbf{Task \& Dataset}} & \multicolumn{4}{c}{\textbf{Performance Scores}} & \multirow{2}{*}{\textbf{Best vs Baseline}} \\
\cmidrule(lr){2-5}
& \textbf{Baseline (0-Shot)} & \textbf{0-Shot} & \textbf{5-Shot} & \textbf{20-Shot} & \\
\midrule
\endhead
\midrule
\multicolumn{6}{r}{\textit{Continued on next page}} \\
\endfoot
\bottomrule
\endlastfoot
\multicolumn{6}{@{}l@{}}{\cellcolor{gray!10}\textbf{\large QUESTION ANSWERING}} \\[0.5ex]
\midrule
\textbf{FinQA} & & & & & \\
\quad $\circ$ Exact Match Acc. & 4.0 & \textcolor{red}{\textbf{7.4}} & 7.0 & -- & \textcolor{green}{+3.4} \\
\addlinespace[0.3ex]
\textbf{ConvFinQA} & & & & & \\
\quad $\circ$ Exact Match Acc. & 20.0 & \textcolor{red}{\textbf{36.4}} & -- & -- & \textcolor{green}{+16.4} \\
\midrule
\multicolumn{6}{@{}l@{}}{\cellcolor{gray!10}\textbf{\large STOCK MOVEMENT PREDICTION}} \\[0.5ex]
\midrule
\textbf{BigData22} & & & & & \\
\quad $\circ$ Accuracy & 53.0 & 50.5 & \textcolor{red}{\textbf{51.5}} & -- & \textcolor{orange}{-1.5} \\
\addlinespace[0.3ex]
\textbf{ACL18} & & & & & \\
\quad $\circ$ Accuracy & 56.0 & 51.2 & 51.1 & -- & \textcolor{orange}{-4.8} \\
\addlinespace[0.3ex]
\textbf{CIKM18} & & & & & \\
\quad $\circ$ Accuracy & 53.0 & 49.3 & 48.7 & -- & \textcolor{orange}{-4.3} \\
\midrule
\multicolumn{6}{@{}l@{}}{\cellcolor{gray!10}\textbf{\large TEXT SUMMARIZATION}} \\[0.5ex]
\midrule
\textbf{EctSumm Dataset} & & & & & \\
\quad $\circ$ ROUGE Score & 8.0 & 2.8 & -- & -- & \textcolor{orange}{-5.2} \\
\bottomrule
\end{longtable}

\begin{figure}[H]
    \centering
    \includegraphics[width=0.8\linewidth]{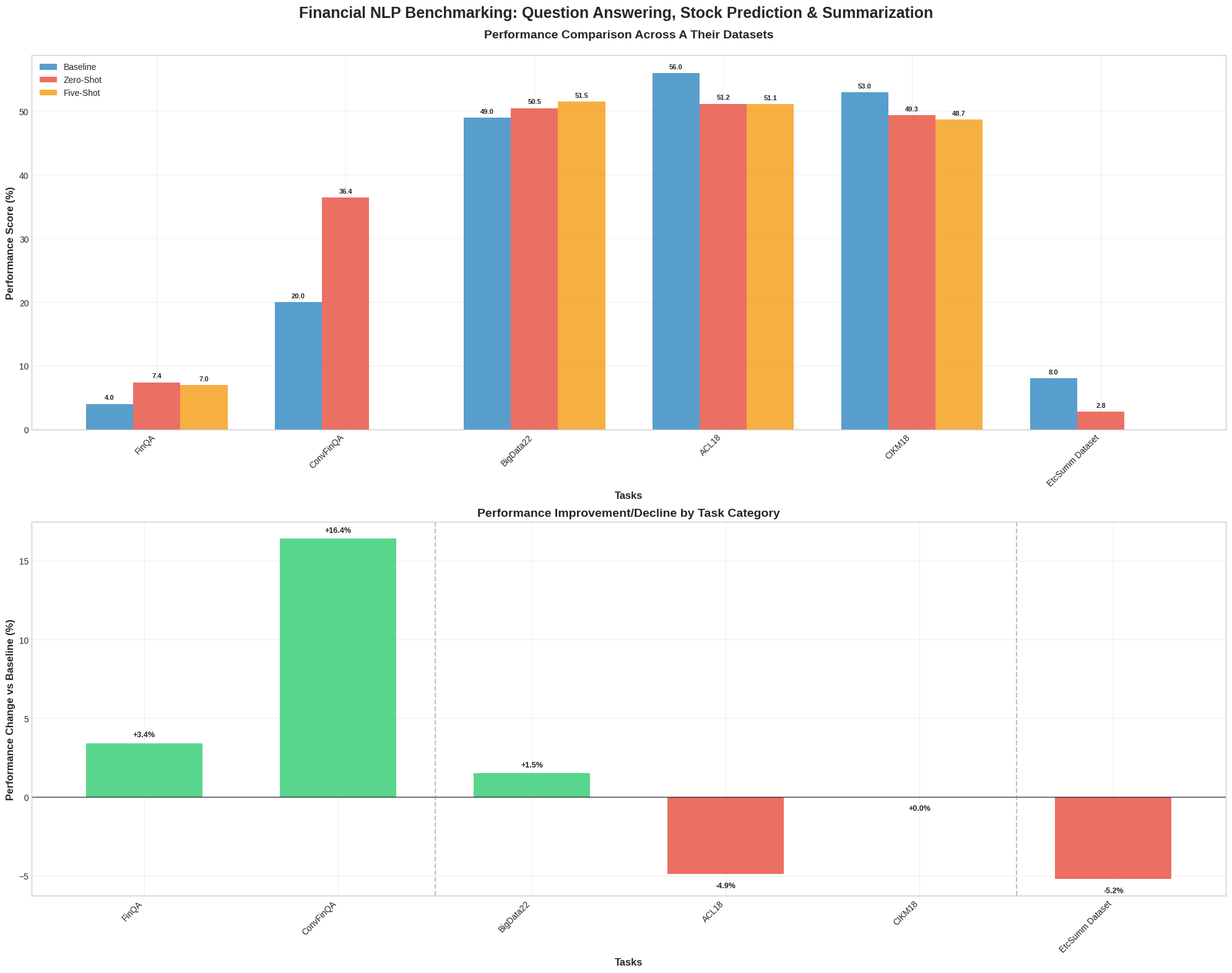}
    \caption{Zero-shot and Few-shot Performance for Question Answering (FinQA, ConvFinQA), Stock Movement Prediction (BigData22, ACL18, CIKM18), and Text Summarization (EctSumm).}
    \label{fig:qa_smp_summ_results}
\end{figure}

\section{Analysis and Interpretation}
\label{sec:analysis}

\textbf{Sentiment Analysis, Headline Classification, and Named Entity Recognition Tasks}

The FinMA-7B-full model demonstrates strong zero-shot performance on sentiment analysis tasks, achieving an F1 score of 93.9\% on FPB and 83.5\% on FiQA-SA, both exceeding the reported baselines of 87.0\% and 79.0\%, respectively (see Table~\ref{fig:sa_tc_ner_results}). These scores also surpass those of FinMA-30B (88\% on FPB) and GPT-4 (93\% on FPB, reported in 5-shot settings) \cite{ Xie2023pixiu}. For news headline classification, FinMA-7B-full achieves an average F1 score of 97.5\% on the Headlines dataset, marginally outperforming its baseline (97.0\%) and significantly surpassing GPT-4 (93\%). However, a slight decline in 5-shot performance (e.g., 82.6\% on FiQA-SA and 93.5\% on Headlines) is observed, which may stem from the model’s sensitivity to example quality, an issue previously identified in LLaMA-based architectures \cite{Touvron2023llama}. These findings are consistent with the instruction tuning methodology adopted from the FIT dataset(\ref{sec:fit_dataset}), and support downstream applications such as sentiment-driven trading or automated news monitoring.

By contrast, results in named entity recognition (NER) are less impressive. In a 5-shot setting, FinMA-7B-full reaches an F1 score of 64.1\%, falling below its own zero-shot baseline of 69.0\%. While it still outperforms BloombergGPT (61\% in 20-shot), it remains far behind GPT-4 (83\%). This underperformance likely stems from LLaMA’s general-purpose architecture, which lacks optimized mechanisms for span detection, as well as the relatively limited representation of financial NER tasks in the FIT training set.

\textbf{Question Answering, Stock Movement Prediction, and Text Summarization Tasks }

In question answering (QA), FinMA-7B-full obtains zero-shot Exact Match scores of 7.4\% on FinQA and 36.4\% on ConvFinQA, improving on the baselines of 4.0\% and 20.0\%, respectively. Although these gains are significant, they remain below GPT-4's 69–76\% and FinMA-30B’s range of 11–40\%. The relatively better performance on ConvFinQA can be linked to conversational formatting in the FIT dataset, which helps the model understand multi-turn questions. However, LLaMA’s known limitations in numerical reasoning, demonstrated by its low accuracy on mathematical benchmarks \cite{Touvron2023llama},likely reduce its effectiveness on finance-specific QA tasks. The absence of 5-shot results on ConvFinQA further restricts the scope of comparative analysis.

For stock movement prediction (SMP), the model shows mixed results. It achieves 50.5\% accuracy on BigData22, 51.2\% on ACL18, and 49.3\% on CIKM18, all below the respective baselines by 1.5 to 4.8 percentage points. Even in 5-shot settings, improvements are marginal. These results illustrate the difficulty the model faces in handling multimodal inputs, such as tweets combined with time series data, which are central to financial forecasting tasks. This limitation affects the model’s applicability in real-world investment strategies.

Finally, the model's performance in text summarization is considerably lower. On the ECTSum dataset, FinMA-7B-full achieves a zero-shot ROUGE-1 score of just 2.8\%, significantly below the 8.0\% baseline and far from GPT-4 (30\%) or specialized summarization models (around 47\%). This gap likely results from the limited presence of summarization instructions in the FIT dataset, which hinders the model’s ability to generate structured and informative financial reports. Moreover, the lack of 5-shot or 20-shot evaluations prevents further insights into whether few-shot prompting could mitigate this shortfall.

\section{Discussion}
\label{sec:discussion}

This section examines FinMA-7B-full’s performance on the FLARE benchmark, highlighting practical implications, challenges, and future directions, drawing on the PIXIU framework \cite{Xie2023pixiu}, based on LLaMA’s architecture \cite{Touvron2023llama}.

\subsection{Practical Implications}
\label{subsec:practical_implications}

FinMA-7B-full excels in zero-shot SA (93.9\% F1 on FPB, 83.5\% on FiQA-SA) and TC (97.5\% average F1 on Headlines), surpassing FinMA-30B (88\% F1), GPT-4 (93\% F1), and BloombergGPT (51--82\% F1). Leveraging LLaMA’s efficient transformer architecture with RMSNorm, SwiGLU activation, and RoPE embeddings \cite{Touvron2023llama}, combined with FIT dataset tuning, enables cost-effective applications in sentiment-driven trading and automated news analysis. The model’s QA performance (36.4\% EM Acc on ConvFinQA) supports financial chatbots and analyst tools, though limitations in numerical reasoning restrict complex reporting tasks \cite{Lee2024survey}. Weaker performance in NER (64.1\% F1), SMP (50.5--51.2\% Acc), and summarization (2.8\% ROUGE) limits applications requiring precise entity detection, multi-modal data integration, or concise report generation.

Ethical considerations include the risk of hallucinations, with LLaMA’s 57\% TruthfulQA score indicating potential misinformation in financial predictions. Biases in LLaMA’s 67\% CommonCrawl training data necessitate expert validation or Retrieval-Augmented Generation (RAG) to ensure reliability. Privacy concerns, particularly with sensitive financial data, require anonymization and adherence to regulatory standards \cite{Touvron2023llama,Lee2024survey}.

\subsection{Challenges}
\label{subsec:challenges}

FinMA-7B-full faces several challenges that limit its performance on the FLARE benchmark, reflecting common issues in financial LLMs. These challenges are supported by evidence from the FIT dataset’s composition (Section~\ref{sec:fit_dataset}, Table~\ref{tab:fit}) and model characteristics.

\begin{enumerate}
    \item \textbf{Limited Financial Training}: FinMA-7B-full relies on LLaMA’s general-purpose 7B-parameter model, which struggles with financial-specific tasks. For named entity recognition (NER), the 5-shot Entity F1 score of 64.1\% (-4.9\% below baseline) is lower than GPT-4 (83\%). This is due to the small size of the FIN Agreements dataset in FIT, with only 13,660 instruction samples (Table~\ref{tab:fit}), limiting the model’s exposure to financial entities like organizations or locations.
    
    \item \textbf{Poor Numerical Skills}: FinMA’s 7.4\% Exact Match accuracy on FinQA (Table~\ref{tab:financial_nlp_results_2}) reflects LLaMA’s weak mathematical reasoning (10.6\% on MATH) \cite{Touvron2023llama}. The FIT dataset’s QA components, FinQA (8,281 samples) and ConvFinQA (3,892 samples), focus more on textual reasoning than numerical calculations (Table~\ref{tab:fit}), reducing FinMA’s ability to handle quantitative financial tasks \cite{Lee2024survey}.
    
    \item \textbf{Multi-Modal Data Issues}: Stock movement prediction (SMP) accuracies range from 50.5\% to 51.2\% (-1.5\% to -4.8\% below baseline) (Table~\ref{tab:financial_nlp_results_2}), below state-of-the-art models (58\%). The FIT dataset’s SMP datasets (BigData22: 7,164 samples; ACL18: 27,053; CIKM18: 4,967) include text and time-series data (Table~\ref{tab:fit}), but LLaMA’s text-only architecture struggles to integrate these modalities, a key challenge in financial prediction \cite{Lee2024survey}.
    
    \item \textbf{Weak Summarization Training}: The 2.8\% ROUGE score on EctSumm is far below top models (47\%). The FIT dataset lacks dedicated summarization tasks (Table~\ref{tab:fit}), with no datasets explicitly designed for financial report summarization, limiting FinMA’s ability to generate concise outputs \cite{Xie2023pixiu}.
    
    \item \textbf{Few-Shot Instability}: 5-shot performance drops, such as TC (93.5\%) and FiQA-SA (82.6\%) (Table~\ref{tab:financial_nlp_results_1}), show LLaMA’s sensitivity to example quality, consistent with variance in social intelligence tasks (SIQA) \cite{Touvron2023llama}. The FIT dataset’s prompt diversity (Section~\ref{sec:fit_dataset}) helps zero-shot performance but does not fully stabilize few-shot learning.
    
    \item \textbf{Limited Computing Power}: Restricted resources prevented full FinMA-30B fine-tuning (Section~\ref{subsec:fine_tuning}), as the FIT dataset’s 136,609 samples required significant computational power (Table~\ref{tab:fit}), impacting scalability \cite{Xie2023pixiu}.
    
    \item \textbf{Bias and Ethical Risks}: Biases in LLaMA’s CommonCrawl training data (67\% of training corpus) may skew financial outputs \cite{Touvron2023llama}. The FIT dataset’s reliance on public sources like tweets and news (Table~\ref{tab:fit}) introduces potential biases, and LLaMA’s 57\% TruthfulQA score highlights risks of incorrect predictions, requiring ethical safeguards \cite{Lee2024survey}.

        \item \textbf{Environment Constraints}: Also for this work, re-evaluation of FinMA-7B-full used Kaggle’s environment with T4 GPUs (16GB, up to 4 CPU cores) and 20GB temporary disk (Table~\ref{tab:experimental_environment}), which falls short of the 8× A100 40GB GPUs used for training (Section~\ref{subsec:fine_tuning}). This limited memory and processing power slowed inference and restricted experiments, such as larger batch sizes or FinMA-30B evaluation, impacting scalability.
\end{enumerate}

Attempts to improve accuracy using prompt engineering techniques, such as Chain-of-Thought or Meta Prompting, did not yield significant gains, particularly for QA (7.4\% FinQA) and SMP (50.5–51.2\%, Table~\ref{tab:financial_nlp_results_2}). The complexity of FIT’s financial tasks, like numerical reasoning in FinQA or multi-modal analysis in BigData22 (Table~\ref{tab:fit}), requires domain-specific prompts, which were limited by a lack of financial expertise in prompt design \cite{Lee2024survey}.

\subsection{Potential Future Directions}
\label{subsec:future_directions}

To address the challenges outlined in Section~\ref{subsec:challenges} above and enhance FinMA-7B-full’s performance within the PIXIU framework \cite{Xie2023pixiu}, several research directions are proposed, drawing on experimental results (Tables~\ref{tab:financial_nlp_results_1}, \ref{tab:financial_nlp_results_2}) and the FIT dataset (Table~\ref{tab:fit}). 

Fine-tuning on specialized financial datasets like FINER-139 for named entity recognition (NER) and EarningsCall for summarization can improve performance, as the FIT dataset’s limited samples (e.g., 13,660 for FIN Agreements) and lack of summarization tasks hinder current results (Table~\ref{tab:fit}). 

Incorporating Retrieval-Augmented Generation (RAG) can further enhance NER accuracy. Training on numerical datasets like GSM8K can strengthen question answering (QA) capabilities, addressing FinMA’s weak 7.4\% Exact Match accuracy on FinQA (Table~\ref{tab:financial_nlp_results_2}) \cite{Touvron2023llama, Lee2024survey}. 

For stock movement prediction (SMP), developing hybrid models to integrate text and time-series data from FIT’s datasets (e.g., BigData22) can boost accuracies (50.5–51.2\%, Table~\ref{tab:financial_nlp_results_2}) \cite{Lee2024survey}.

Active learning to select high-quality examples can stabilize few-shot performance in tasks like news headline classification (TC: 93.5\%, Table~\ref{tab:financial_nlp_results_1}) \cite{Touvron2023llama}. The lack of accuracy gains from prompt engineering techniques, such as Chain-of-Thought, suggests a need for financial domain expertise to design tailored prompts for FIT’s complex tasks (Table~\ref{tab:fit}) \cite{Lee2024survey, Chen2024}. Transitioning from Kaggle’s T4 GPUs (Table~\ref{tab:experimental_environment}) to cloud platforms like AWS can enable faster inference and FinMA-30B experiments. 

\chapter{Conclusion}
\label{chap:conclusion_generale}

This work evaluated the performance of \textbf{FinMA-7B-full} model on the FLARE benchmark from the PIXIU framework \cite{Xie2023pixiu}, as detailed in Chapter~\ref{chap:results_analysis}. The results show significant progress in sentiment analysis (SA)(93.9\% F1 on FPB, +6.9\% compared to the baseline) and news headline classification (TC) (97.5\% Avg F1 on Headlines, +0.5\% over the baseline), outperforming a little bit FinMA-30B and GPT-4 in zero-shot settings. These achievements, supported by the efficient LLaMA architecture \cite{Touvron2023llama} and fine-tuning on the FIT dataset, confirm the potential of FinMA-7B-full for financial applications such as sentiment-based trading and news analysis \cite{Lee2024survey, Nie2024}.

However, limitations remain in named entity recognition (NER) task (64.1\% F1, -4.8\% compared to the baseline), stock movement prediction (SMP) (50.5--53.0\% accuracy), and summarization generation (Summ) (2.8\% ROUGE). These issues are mainly due to the general-purpose nature of the LLaMA backbone and limited financial training \cite{Touvron2023llama, Xie2023pixiu}. Biases in the LLaMA CommonCrawl data and hallucination risks (57\% TruthfulQA) highlight the need for expert validation and ethical measures such as differential privacy \cite{Nie2024, Lee2024survey}.

The main contributions of this work include a rigorous evaluation of FinMA-7B-full, highlighting its strengths in financial language understanding and its weaknesses in numerical reasoning and multimodal tasks. These findings enrich the literature on FinLLMs, building on works like PIXIU \cite{Xie2023pixiu} and BloombergGPT, and offer perspectives for practical applications in finance.

\textbf{Future Work:} Future research will focus on the inference and evaluation of advanced conversational AI models applied to financial tasks. Although GPT-4.5 was initially considered due to its enhanced reasoning capabilities and reduced hallucination rates, its high inference cost and planned deprecation\footnote{https://medium.com/artificial-synapse-media/openai-deprecates-gpt-4-5-api-in-july-2025-forcing-developers-to-migrate-to-gpt-4-1-amid-backlash-417a4a31eb0d} present notable limitations for long-term use. Consequently, future research will consider more practical and cost-effective alternatives such as GPT4.1 and GPT4-turbo (o4), which represent the most capable models publicly accessible at the time of writing, along with open-source language models fine-tuned for the financial domain, including FinMA-7B-full. Model performance will be evaluated using selected tasks from the FLARE benchmark \cite{Xie2023pixiu}, with a particular focus on sentiment analysis, question answering, and multimodal prediction. 

For fine-tuning FinMA-7B-full, the first priority after securing access to suitable GPU resources will be to explore Low-Rank Adaptation (LoRA) techniques \cite{hu2021lora}. LoRA enables efficient and cost-effective fine-tuning by updating only a small subset of model parameters, significantly reducing GPU memory requirements and training time compared to full fine tuning. This approach is particularly suitable for large models like FinMA-7B and facilitates experimentation even on limited hardware. We will also investigate the potential merging or knowledge transfer between FinMA and FinGPT models to leverage complementary strengths. Additionally, the implementation of retrieval-augmented generation (RAG) methods will be considered to reduce hallucinations and enhance factual consistency in financial AI applications.

\newpage

\bibliographystyle{plainnat}
\bibliography{references}

\end{document}